\documentclass[preprint]{ieeeaccess}

\usepackage{cite}
\usepackage[utf8]{inputenc} 
\usepackage[T1]{fontenc}    
\usepackage{hyperref}       
\usepackage{url}            
\usepackage{booktabs}       
\usepackage{amsfonts}       
\usepackage{nicefrac}       
\usepackage{microtype}      
\usepackage[utf8]{inputenc} 
\usepackage{float} 
\usepackage{array} 
\usepackage{graphicx}
\usepackage{xcolor} 
\usepackage{colortbl} 
\usepackage{caption} 
\usepackage{booktabs} 
\usepackage{adjustbox} 
\usepackage{newunicodechar}
\usepackage{subcaption}
\usepackage{makecell}
\usepackage{pifont} 

\usepackage{fontawesome}

\usepackage{amsmath}
\usepackage{graphicx}
\usepackage{subcaption}

\usepackage{multirow}
\newcolumntype{P}[1]{>{\centering\arraybackslash}p{#1}}

\definecolor{LightGrey}{rgb}{0.9,0.9,0.9}

\newcolumntype{R}[1]{>{\centering\arraybackslash}p{#1}}

\def\BibTeX{{\rm B\kern-.05em{\sc i\kern-.025em b}\kern-.08em
    T\kern-.1667em\lower.7ex\hbox{E}\kern-.125emX}}

\usepackage{newunicodechar}
\newunicodechar{✓}{\ding{51}}
\newunicodechar{✗}{\ding{55}}

\begin{document}

\title{EgoWalk: A Multimodal Dataset for Robot Navigation in the Wild}
\author{
\uppercase{Timur Akhtyamov}\authorrefmark{1,*},
\uppercase{Mohamad Al Mdfaa \authorrefmark{1,*}}
\uppercase{Javier Antonio Ramirez Benavides \authorrefmark{1}}
\uppercase{Arthur Nigmatzyanov \authorrefmark{1}}
\uppercase{Sergey Bakulin \authorrefmark{1}}
\uppercase{German Devchich \authorrefmark{1}}
\uppercase{Denis Fatykhov  \authorrefmark{1}}
\uppercase{Diego Ruiz Salinas \authorrefmark{1}}
\uppercase{Alexander Mazurov \authorrefmark{1}}
\uppercase{Kristina Zipa \authorrefmark{1}}
\uppercase{Malik Mohrat \authorrefmark{2}}
\uppercase{Pavel Kolesnik  \authorrefmark{2}}
\uppercase{Ivan Sosin   \authorrefmark{2}}, and
\uppercase{Gonzalo Ferrer  \authorrefmark{1}}
.
}
\address[1]{Skolkovo Institute of Science and Technology, 121205 Skolkovo, Russia}
\address[2]{Sber Robotics Center, 121165 Moscow, Russia}
\address[*]{Shared first author}


\markboth
{Akhtyamov \headeretal: EgoWalk: A Multimodal Dataset for Robot Navigation in the Wild}
{Akhtyamov \headeretal: EgoWalk: A Multimodal Dataset for Robot Navigation in the Wild}

\corresp{Corresponding author: Timur Akhtyamov (e-mail: timur.akhtyamov@skoltech.ru).}

\begin{abstract}

Data-driven navigation algorithms are critically dependent on large-scale, high-quality real-world data collection for successful training and robust performance in realistic and uncontrolled conditions.
To enhance the growing family of navigation-related real-world datasets, we introduce EgoWalk — a dataset of 50 hours of human navigation in a diverse set of indoor/outdoor, varied seasons, and location environments. Along with the raw and Imitation Learning-ready data, we introduce several pipelines to automatically create subsidiary datasets for other navigation-related tasks, namely natural language goal annotations and traversability segmentation masks. Diversity studies, use cases, and benchmarks for the proposed dataset are provided to demonstrate its practical applicability.

We openly release all data processing pipelines and the description of the hardware platform used for data collection to support future research and development in robot navigation systems.
\end{abstract}


\titlepgskip=-15pt

\maketitle

\section{Introduction}


\begin{figure*}[h]
  \centering
  \includegraphics[scale=0.3]{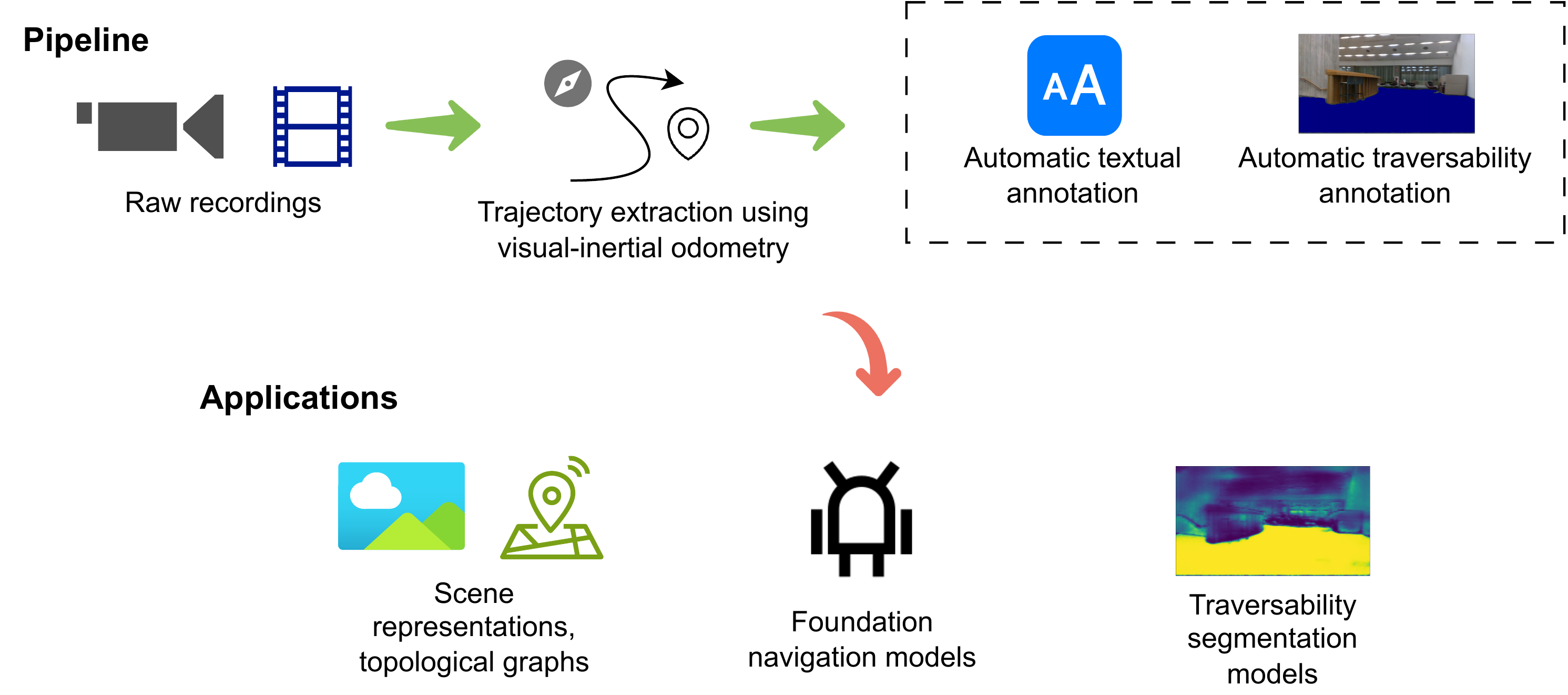}
  \caption{General overview of the data collection and processing pipelines. Sensor and odometry data are extracted from 50 hours of egocentric recordings and can be directly used for general navigation-related tasks. An automatic traversability region and language goals annotation pipeline are introduced to enlarge the scope of potential applications.
  }
  \label{fig:visual_abstract}
\end{figure*}

While robots achieve lower trajectory error in controlled settings, humans consistently attain higher task success rates and better adherence to social navigation norms, particularly when generalizing to previously unseen environments \cite{riener2023robots, akhtyamov2025social}.
Intuitively, humans' capabilities of employing experience, understanding environment semantics, and linguistic context make a significant contribution to this phenomenon. This principle made imitation learning (IL), visual, vision-language, and semantics-aware navigation one of the main research directions in robotics. Despite their advantages and prospects, one of the main limitations that they have in common is the need for a large amount of high-quality real-world data \cite{black2410pi0}.

Specifically, IL today has become a dominant paradigm for various branches of robotics, such as manipulation \cite{kim2021transformer, xie2020deep}, navigation \cite{shah2023gnm, shah2023vint, sridhar2024nomad}, and locomotion \cite{doshi2024scaling, tang2024humanmimic}. Various large-scale datasets can be found for the manipulation task \cite{vuong2023open, walke2023bridgedata, khazatsky2024droid}. However, for the navigation task, the amount of high-quality annotated real-world data is limited. Recent works proposed data mining strategies from YouTube videos \cite{hirose2024lelan, liu2024citywalker}, which allow a significant increase in the number of navigation hours but lack ground truth metric trajectories and are not suitable for multi-sensor scenarios. Although scene understanding, as outlined above, is crucial for real-world navigation, available datasets generally view actual navigation \cite{T8/0PRYRH_2022, nguyen2023toward} and scene semantics \cite{10943903} as separate tasks.

To close the outlined gaps, we introduce \textit{EgoWalk} -- a novel egocentric navigation dataset of more than 50 hours of real-world navigation data, collected with an industry-grade stereo camera in a diverse set of places and conditions. Inspired by an IL-based navigation task, it is structured to naturally support navigation and semantics approaches beyond IL, such as topological mapping, scene understanding, representation learning, traversability estimation, and natural language-based navigation. In particular, we introduce two semantics-aware use cases: automatic traversability mask generation and natural language goals annotation.

\begin{figure*}
    \centering
    \begin{subfigure}[b]{0.35\linewidth}
            \includegraphics[height=3.5cm, width=0.75\linewidth]{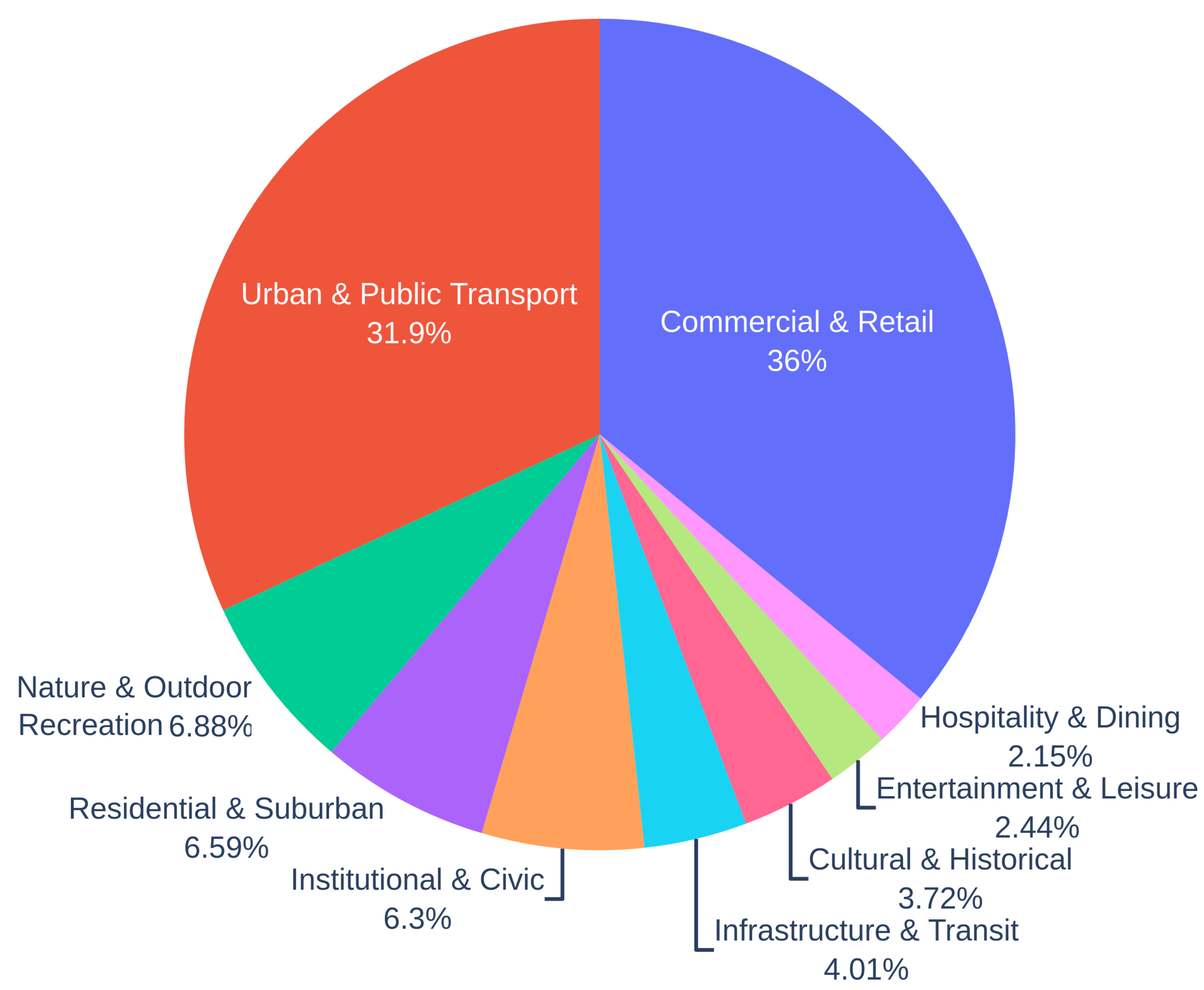}
            \caption{Statistics over locations}
            \label{fig:diversity_stats_places}
    \end{subfigure}
    \begin{subfigure}[b]{0.45\linewidth}
            \centering
            \includegraphics[height=3.2cm, width=0.75\linewidth]{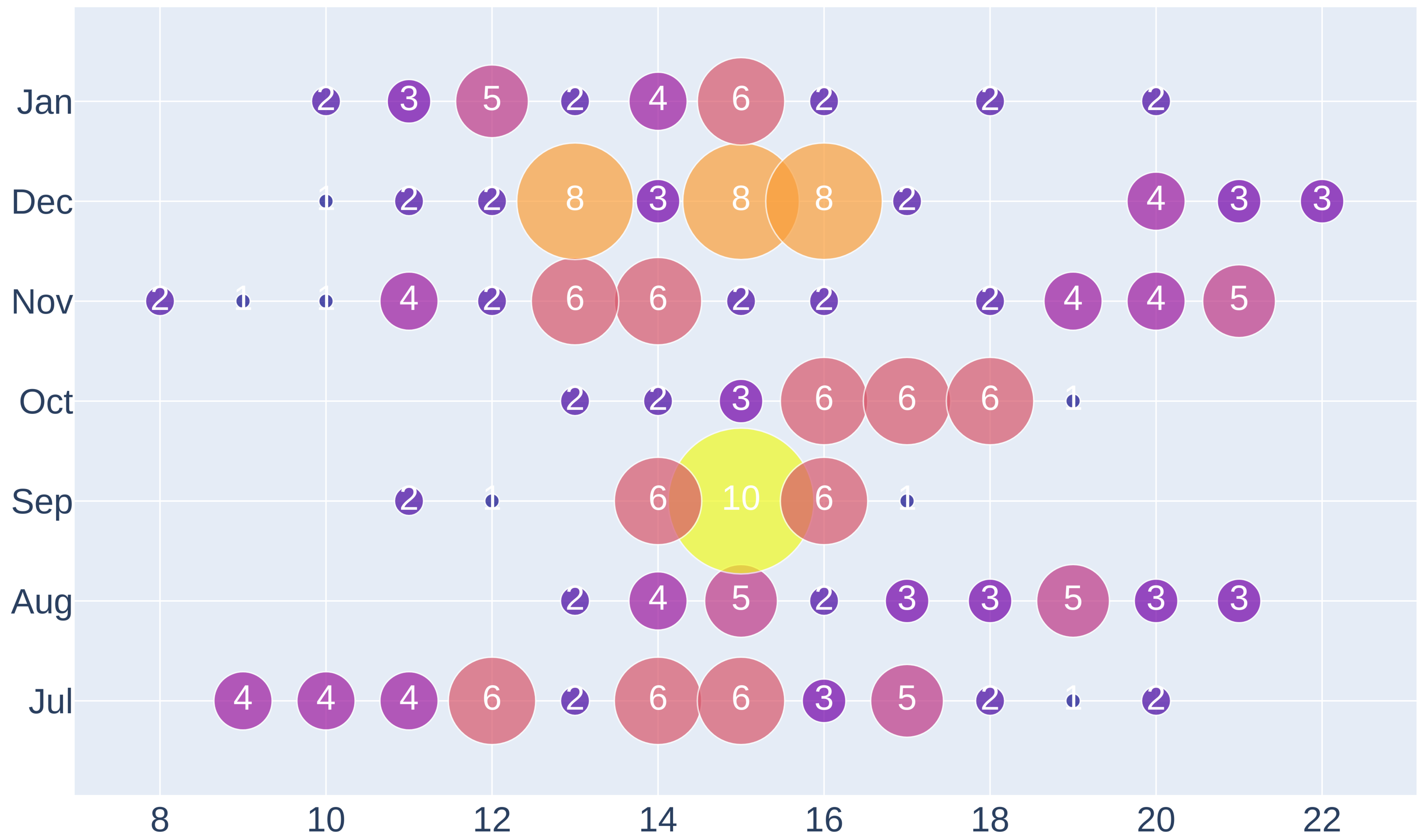}
            \caption{Statistics on seasons and times of day}
            \label{fig:diversity_stats_time}
    \end{subfigure}
    \caption{Diversity of the dataset. Location labels were produced using a vision-language model \cite{hong2024cogvlm2}.}
    \label{fig:diversity_stats}
\end{figure*}

Figure \ref{fig:visual_abstract} summarizes our contributions. We publicly release a large-scale dataset in several forms: source raw recordings, extracted odometry-paired trajectories, and two subsidiary datasets with traversability masks and natural language annotations, respectively. We demonstrate the practical applicability of EgoWalk to train a visual navigation policy for a real robot and several traversability prediction models. Additionally, we show its applicability for scene reconstruction, pose estimation, and language-based annotation tasks. All processing pipelines are open-source, along with the design and software of the data recording platform.

All required data, code, and documentation links can be found on the paper's website. \footnote[1]{\url{https://sites.google.com/view/egowalk}}
\footnotetext{This work has been submitted to the IEEE for possible publication. Copyright may be transferred without notice, after which this version may no longer be accessible.}

\section{Related Works} \label{sec:related_works}

Before discussing the collected data, we first review tasks and datasets to provide the actual motivation to collect the new dataset.

\textbf{Visual navigation and related approaches.} Visual navigation (VN) of the robot can be formulated as a collision-free movement of the robot relying solely on the RGB (vision-only navigation) or sometimes on the RGB-D signal \cite{bonin2008visual, yasuda2020autonomous, zhang2022survey}. This includes both goal-free navigation (exploration \cite{sridhar2024nomad}), visual or coordinates goal conditioning \cite{shah2023vint, shah2022viking}, and natural language goal or instruction conditioning known as Vision-Language Navigation (VLN) problem \cite{park2023visual, wu2024vision, gu2022vision}. Unlike classical sensor-rich SLAM and planning stacks, VN and VLN are still challenging problems, especially for the RGB-only case. Early works used classical computer vision concepts \cite{ohnishi2008visual, mccarthy2004performance}. With the advancement of Deep Learning (DL), many Deep Reinforcement Learning (DRL) methods have emerged \cite{zeng2020survey, kulhanek2021visual}. However, the simulation-to-reality (sim2real) problem is the main challenge for the DRL-based approaches. A recent trend that allows to tackle sim2real issue is the use of Imitation Learning (IL) with real-world datasets \cite{brohan2022rt, shah2023gnm, shah2023vint, sridhar2024nomad, doshi2024scaling, liu2024citywalker, gode2024flownav}. These IL-based approaches became a huge step towards universal real-world-enabled policies. However, their performance highly depends on the amount and quality of the data.


\textbf{Semantics in navigation.} The task-specific semantics aspect of the scene is present in classical, sensor-rich navigation. Despite the availability of precise localization and obstacle detection, the final maneuver is found to be highly dependent on features that are rarely presented in dense maps (traversability \cite{jung2024v, kim2024learning}, pedestrian detection \cite{hirose2023sacson}, and panoptic information \cite{10943903}). Typically, RGB is the only modality employed for the extraction of such features, underscoring the significance of semantics-aware navigation datasets.

From the visual and semantics-aware navigation point of view, we introduce the dataset criteria to train generalizable and agile navigation agents and/or their components:
\begin{itemize}
    \item Scale in terms of data size (e.g. in navigation hours) due to the requirements of IL models \cite{shah2023vint, doshi2024scaling};
    \item Diversity in terms of location types, weather conditions, time of the day, etc., which naturally follows from the nature of IL models;
    \item Availability of the task-specific semantics and/or a way to produce and align them with the classical range-based navigation.
\end{itemize}

\begin{table*}
  \scriptsize 
  \setlength{\tabcolsep}{3pt} 
  \caption{Comparison of navigation datasets. The question mark indicates inability to assess due to the large dataset size.}
  \label{tab:datasets_cmp}
  \centering
  \begin{tabular}{cccccccc} 
    \toprule
    \textbf{Dataset} & \textbf{\makecell{Data \\ source}} & \textbf{\makecell{Duration \\ (hours)}} & \textbf{\makecell{Indoor/ \\ Outdoor}} & \textbf{\makecell{Range \\ Sensing}}       & \textbf{\makecell{Language \\ Annotations}} & \textbf{\makecell{Semantic \\ Annotations}} & \textbf{\makecell{All- \\ weather}} \\
    \midrule
    SCAND \cite{T8/0PRYRH_2022}            & \makecell{Teleoperated \\ robot} & 8.7                        & \ding{51} / \ding{51}   & \makecell{3D LiDAR, \\ RGBD/ stereo camera} & \ding{55}                            & \makecell{Social \\ interaction tags}       & \ding{55} \\ 
    MuSoHu \cite{nguyen2023toward}          & Human                   & 20                         & \ding{51} / \ding{51}   & \makecell{3D LiDAR, \\ stereo camera}      & \ding{55}                            & \makecell{Social \\ interaction tags}       & \ding{55} \\ 
    SACSoN \cite{hirose2023sacson}          & \makecell{Autonomous \\ robot} & 75                         & \ding{51} / \ding{55}   & \makecell{2D LiDAR, \\ spherical RGBD}     & \ding{55}                            & \makecell{People detections}             & \ding{55} \\ 
    SANPO \cite{10943903}           & Human                   & 14.5                       & \ding{55} / \ding{51}   & Stereo cameras                             & \ding{55}                            & \makecell{Panoptic \\ segmentation}         & \ding{51} \\ 
    LeLaN \cite{hirose2024lelan}           & YouTube                 & 130                        & \ding{51} / \ding{51}   & \ding{55}                                  & \makecell{Sparse \\ navigation goals}       & \ding{55}                                  & ? \\ 
    CityWalker \cite{liu2024citywalker} & YouTube                 & 2000+                      & \ding{51} / \ding{51}   & \ding{55}                                  & \ding{55}       & \ding{55}                                  & ? \\ 
    RELLIS-3D \cite{jiang2021rellis} & \makecell{Autonomous \\ robot} & 0.5 & \ding{55} / \ding{51} & \makecell{3D LiDAR, \\ RGB, stereo camera} & \ding{55} & \ding{51} & \ding{55}  \\
    EgoWalk (ours)   & Human                   & 50                         & \ding{51} / \ding{51}   & Stereo camera                              & \makecell{Sparse \\ navigation goals}       & \makecell{Sparse \\ traversability \\ masks}   & \ding{51} \\ 
    \bottomrule
  \end{tabular}
\end{table*}
\textbf{Navigation Datasets.} An overview of the main large real-world navigation-related datasets is provided in Table \ref{tab:datasets_cmp}. In this comparison we include datasets suitable for the VN-based task, implying the availability of egocentric views and odometry. The table shows that the existing datasets satisfy the requirements outlined above only partially. YouTube-based datasets offer significantly longer durations (130+ hours by \cite{hirose2024lelan} and 2000+ hours by \cite{liu2024citywalker}). However, they cannot provide range data and metric odometry, which are critical components for embodied navigation tasks. EgoWalk aims to satisfy the requirements and provide a trade-off between duration, environmental diversity, and annotation.

\textbf{Automatic Navigation Data Annotation Pipelines.} Recent advances in foundation models, LLMs, and VLMs dramatically reduce the cost of data processing by enabling automatic data annotation pipelines \cite{hirose2024lelan, kim2024learning, jung2024v, yang2024generalized}. We make EgoWalk useful for tasks beyond navigation by providing a sparse (i.e. selected key frames) annotation pipeline for last-mile navigation goals inspired by  \cite{hirose2024lelan} and traversability masks based on \cite{kim2024learning}.

\textbf{Real2Sim paradigm.} Recent advances in 3D reconstruction \cite{wang2025vggt, keetha2026mapanything, lin2025depth}, novel view synthesis (NVS) \cite{mildenhall2021nerf, kerbl20233d}, and world models \cite{koh2021pathdreamer, agarwal2025cosmos, bar2025navigation} have inspired an alternative to standard IL for bridging the sim2real gap. This approach introduces a novel method for building simulation environments: rather than relying on classical physics and manual computer graphics, it employs data-driven reconstruction and rendering techniques grounded in real-world data. This concept is widely known as real2sim transfer \cite{da2025survey}. Recent literature \cite{xie2025vid2sim, vrrobo, he2025seeing} demonstrates the successful application of this paradigm for the training and evaluation of autonomous navigation systems in urban environments. Further works, such as UrbanVerse \cite{liu2026urbanverse}, have highlighted the scalability of anchoring classical simulation-based photo-realistic virtual worlds to real-world touring videos. To support and advance these methodologies, our EgoWalk dataset serves as a novel data source. We provide a corresponding demonstration of EgoWalk applied to 3D reconstruction and 3D Gaussian Splatting.

\section{Data Collection Overview}

This section provides a brief overview of the collected dataset and how it was recorded and organized.

\subsection{Dataset Overview}

The EgoWalk dataset aims to close the gap between the requirements outlined in the section \ref{sec:related_works}, and provides a complete set of human examples to navigate successfully in all common daily environments. A total of 50 hours of data were recorded in Moscow from July 2024 to February 2025. Figures \ref{fig:diversity_stats_time} and \ref{fig:diversity_stats_places} provide visual evidence supporting the dataset's diversity characteristics. EgoWalk covers all major and reasonable times of the day and 3 seasons, which contrast significantly in this geographical location. The diversity of locations reflects typical urban scenarios while also considering less common and more specific environments.


The main data includes 5 FPS trajectories consisting of RGB and depth images and odometry. Additionally, there are language annotations, traversability data and raw data; see the next sections for more details.

\subsection{Hardware Platform Overview}
\label{sec:hardware_platform_overview}

\begin{figure*}[t]
    \centering
    \begin{subfigure}[b]{0.3\linewidth}
        \centering
        \includegraphics[height = 4cm, width=\linewidth, keepaspectratio]{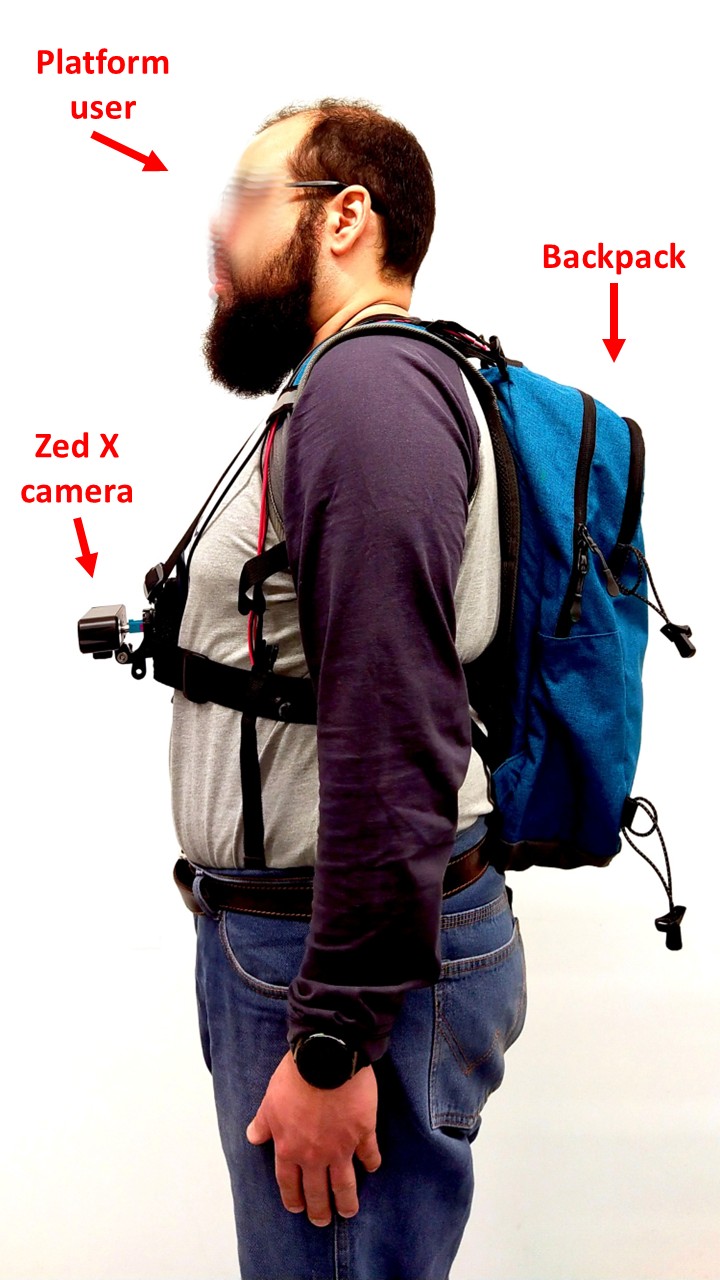}
    \end{subfigure}
    \begin{subfigure}[b]{0.6\linewidth}
        \centering
        \includegraphics[height = 4cm, width=\linewidth, keepaspectratio]{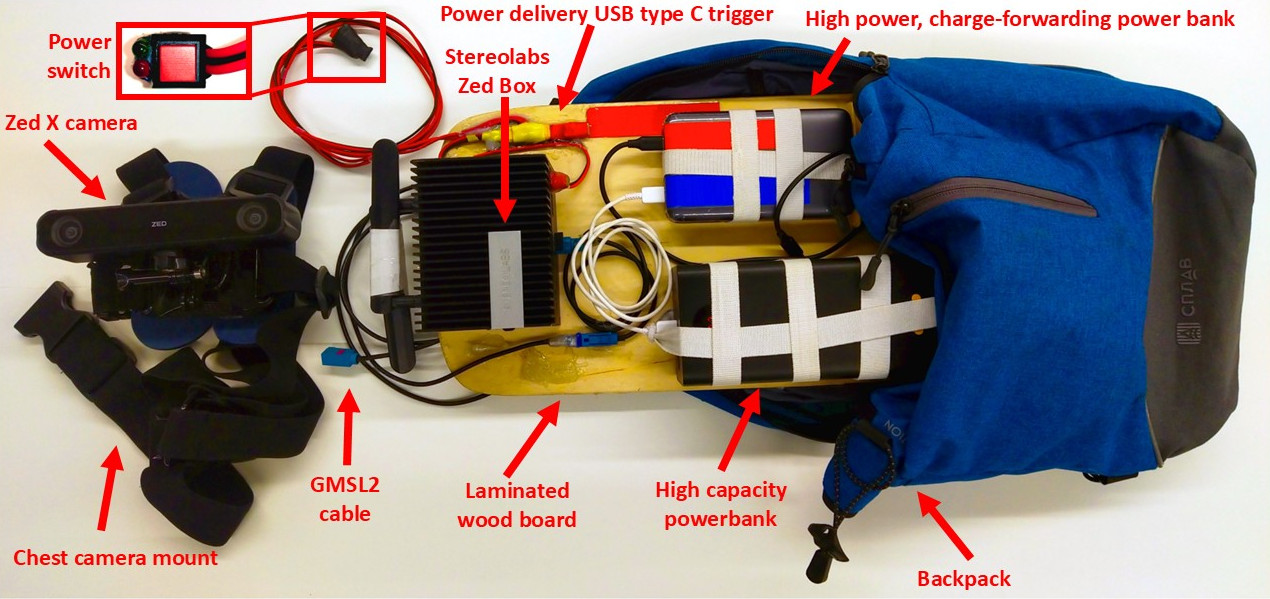}
    \end{subfigure}
    \caption{Data collection platform: an example of the wearable rig (left) and detailed internal component layout (right).}
    
    \label{fig:platform}
\end{figure*}

Our platform is highly inspired by the rig proposed by \cite{10943903}. The setup is demonstrated in Fig. \ref{fig:platform}. The core of the platform is a chest-mounted \textit{ZED 2} stereo camera and \textit{Nvidia Jetson}-backed \textit{ZED Box} compute module. The platform is powered by a system of two power banks. The setup fits a standard backpack; to prevent potential overheating, we asked participants to keep it slightly open when possible. The recording is performed using standard \textit{ZED SDK} tools, and a smartphone is used to control and monitor the process. Bill of materials, instructions, and code links are available at the paper's website.

The raw recordings are stored in the \textit{.svo2} file format supported by the \textit{ZED SDK}. The recording is performed at 30 FPS with SVGA resolution. These data are processed offline using \textit{ZED SDK}, which includes RGB frame extraction, depth, and odometry computation. 

\subsection{Recording Organization and Data Processing} \label{sec:organization_processing}

The data has been recorded by multiple participants (both volunteers and paid employees) using a setup (Figure \ref{fig:platform}) with chest-mounted \textit{ZED X} stereo camera inspired by \cite{10943903}. For all participants, an approximate camera height above the floor is measured to later compute the projection of footsteps \cite{kim2024learning}. Participants were asked to follow the guidelines:
\begin{itemize}
    \item \textit{Robot Centricity}: keeping in mind that their trajectories will be used for robot learning, thus, participants should avoid maneuvers infeasible for a mid-size mobile robot;
    \item \textit{Social interactions}: participants should enable various socially acceptable maneuvers whenever it is possible and logical;
    \item \textit{Collision avoidance}: participants should record collision avoidance examples when it is reasonable and feasible;
    \item \textit{Turn Prioritization}: dominance of linear maneuvers is inevitable when recording normal human motion; to reduce this dominance, participants should prioritize the turns whenever it is more or less logical and feasible.
\end{itemize}
With those instructions, our goal was to make our dataset well suited for robot learning.

For the raw 30 FPS recordings produced by \textit{ZED SDK}, per-frame depth images and odometry poses are calculated. Afterwards, the data rate is reduced to 5 FPS, which matches the rates of common high-level navigation policies \cite{shah2023gnm, doshi2024scaling}. A face blur was applied to each RGB frame to preserve the privacy and personal data of surrounding pedestrians. The OWL-ViT model \cite{minderer2022simple} was used here for face detection. Having carefully collected and preprocessed the data, we proceed to annotation.

\section{Annotation Pipelines}

In this section, we discuss the automatic annotation pipelines that were applied to the main dataset (i.e. 5 FPS). The data preparation for vision-only navigation is not addressed since it is available out of the box after extraction of odometry.

\subsection{Language Annotations}
\label{sec:lang_annotation}

To support the development of modern multimodal navigation models, we provide two complementary language-annotation pipelines. The first
relies on open-vocabulary object detection and is referred to as the \textit{Goal-Boxes} pipeline; the second leverages a large vision-language model (VLM) \cite{google2025gemini3} directly and is referred to as the \textit{End-to-End} pipeline. 
Both pipelines are designed to mine training data for the last-mile navigation task \cite{hirose2024lelan}, and share the following formulation. 
Given a reference frame, a valid navigation goal is a visible object in that frame. The pipeline must: (i) extract a feasible trajectory from the reference pose to the goal, suitable for a local planner, and (ii) produce a textual description of the goal. The two pipelines differ in the paradigm used to solve this problem.

\subsubsection{Goal-Boxes Pipeline}
\label{sec:lang_goal_boxes}

\begin{figure*}[t]
  \centering
  \includegraphics[scale=0.45]{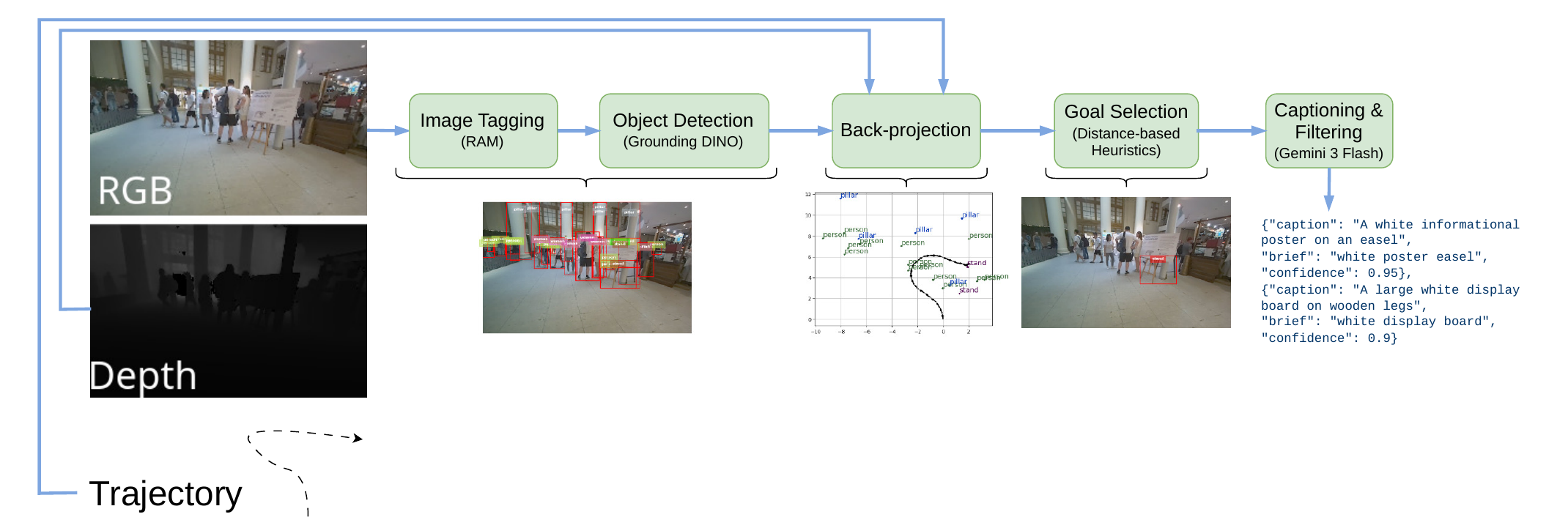}
  \caption{Overview of the \textit{Goal-Boxes} natural-language goal annotation
  pipeline.}
  \label{fig:language_annotation_pipeline}
\end{figure*}

The \textit{Goal-Boxes} pipeline is inspired
by \cite{hirose2024lelan}, which employs a learned navigation
policy \cite{sridhar2024nomad} to plan a trajectory toward a selected target crop. In order to preserve the metric, real-world trajectories presented in EgoWalk, we address the inverse problem. Given a ground-truth expert trajectory, we heuristically select the goal object that best explains it. An overview of the pipeline is shown in Figure \ref{fig:language_annotation_pipeline}.

For a reference RGB frame, we first detect candidate goal objects.
Following recent open-vocabulary perception
pipelines \cite{gu2024conceptgraphs, mdfaa2024mapping}, we combine
RAM \cite{zhang2024recognize} with Grounding
DINO \cite{liu2024grounding}. Compared with the more common
SAM \cite{kirillov2023segment} and CLIP \cite{radford2021learning}
combination, this choice yields semantically more consistent
detections and substantially fewer small or noisy segments that would otherwise be hard to filter out. Using the corresponding metric depth map, we back-project the center of each candidate bounding box into the local metric frame of the reference camera. The future bird's-eye-view (BEV) trajectory of the agent is computed in the same frame from odometry. Candidate objects are then filtered according to the following criteria:
\begin{itemize}
    \item \textit{RAM tag.} Ambiguous and non-informative tags
    are excluded via a hand-curated blacklist.
    \item \textit{Bounding-box size.} Boxes that are too small or too large (e.g., scene-scale tags produced by RAM) are removed.
    \item \textit{Distance from the reference pose.} Objects that are too close to the camera are discarded, as they yield very short and uninformative trajectories. Objects that are
    too far from the trajectory are also discarded, so that the retained goals can be plausibly associated with the executed motion.
\end{itemize}
Among the remaining candidates, the object closest to the trajectory is selected as the navigation goal.

The bounding box of the selected goal, expanded with a small padding, is cropped from the reference frame and passed to a large VLM --- Gemini 3 Flash \cite{google2025gemini3} --- for captioning. We prompt the model to produce two types of captions: a brief one, consisting of the main noun phrase and a small number of adjectives (typically 2--3 words), and a more detailed one, in the form of a richer sentence describing the object's appearance and salient attributes.
For each pair, we ask the model to additionally generate two or three synonymous variants in order to increase annotation diversity. The VLM is also instructed to skip the object if it is uncertain whether the candidate is a distinctive navigation goal. In total, the \textit{Goal-Boxes} pipeline produced 50{,}019 captions.

\subsubsection{End-to-End Pipeline}

The second pipeline exploits the long-context video understanding
capabilities of recent VLMs. Each EgoWalk recording is split into
chunks of 11 seconds (selected heuristically as a trade-off between last-mile setting and VLM query cost), separated by short gaps to encourage diversity
across chunks. From each chunk we assemble a short clip sampled at
1 FPS and pass it to Gemini 3 Flash \cite{google2025gemini3}. The clip is supplied as a sequence of
images annotated with frame indices rather than as a video stream.
The VLM is prompted to (i) identify suitable goal candidates within the clip, if any; (ii) generate brief and detailed captions with synonyms, following the format described in
Section \ref{sec:lang_goal_boxes}; and (iii) report the frame index at which the object first leaves the visible area. 
In total, the \textit{End-to-End} pipeline produced 31{,}504
captions.

\subsection{Traversability Annotations}
Our traversability annotation pipeline is inspired by \cite{kim2024learning}. For a given frame, we project future BEV odometry onto the image as approximate footstep locations. This projection relies on two assumptions: (1) the ground plane is orthogonal to the camera plane, and (2) the camera height is approximately known (see Section \ref{sec:organization_processing}). Assuming that people are walking only within the well-traversed regions, those projected points are used as a prompt for the SAM model to generate masks for these regions.

\begin{figure*}[h]
  \centering
  \includegraphics[scale=0.38]{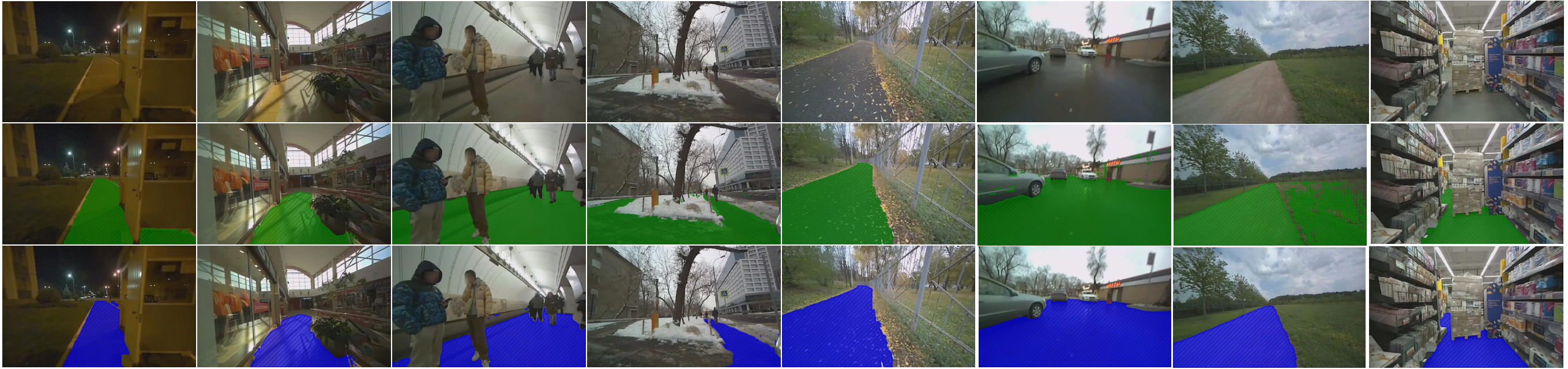}
  \caption{Examples of the auto-generated traversability masks. \textit{Top row:} RGB input images. \textit{Middle row:} traversable masks selected by largest area. \textit{Bottom row:} traversable masks selected by highest score.}
  \label{fig:traverse_gt_examples}
\end{figure*}

The SAM model produces three scored masks for each input image and prompt. We observed that the highest-confidence mask is not always the most suitable for our task. Since the selection remains a heuristic process, we store both the top-scoring mask and the mask with the largest area for each image. Figure \ref{fig:traverse_gt_examples} provides an example set of masks obtained by these criteria. This traversability dataset is produced from the main dataset and released as a separate one, including more than 30,000 entities.

\section{Use Cases and Experiments} 
\label{sec:experiments}
In order to show the applicability of the introduced dataset, we provide several case studies related to the main target tasks outlined in previous sections.
\subsection{Real-World Vision-Only Navigation}
\label{sec:vn_real_world_eval}

To demonstrate the suitability of EgoWalk for
robot navigation, we fine-tune two prominent visual navigation foundation models, ViNT~\cite{shah2023vint} and NoMaD~\cite{sridhar2024nomad}, on EgoWalk and evaluate the resulting policies on a real robot. We rely on the official implementations and training instructions, introducing only the EgoWalk-specific pre-processing required by each model. The official
checkpoints serve as initialization, and EgoWalk is the sole source of \textit{fine-tuning} data. Our objective is to assess how the additional data provided by EgoWalk affects the performance of these state-of-the-art policies.

The fine-tuned models are deployed on a custom mobile platform built on the \textit{AgileX Tracer} chassis. The platform carries an \textit{Intel NUC11PHKI7C000} compute module equipped with a laptop-grade
\textit{NVIDIA RTX 2060} GPU, and uses an \textit{Azure Kinect} camera in HD RGB-only mode as its sole exteroceptive sensor. 
The camera model, intrinsics, the camera mounting height, and the robot's typical motion profile all differ substantially from those of the data in EgoWalk, inducing a non-trivial domain shift. Evaluation experiments were conducted in three indoor environments on the university campus, \textit{none of which are present in the dataset}. 
Topological graphs (same for all models) of these environments were recorded approximately 24 hours prior to the experiments in order to probe the robustness of the policies to short-term scene changes. 
Each model was evaluated over three trials per environment, and the reported success rate is the average across the resulting nine runs. A trial is counted as successful if the robot reaches the target node of the topological graph without collisions. 
The results are summarized in Table~\ref{tab:real-world-nav}.

\begin{table}[t]
\centering
\caption{Real-world vision-only navigation results. Success rate is averaged over three trials in each of three previously unseen environments.}
\label{tab:real-world-nav}
\begin{tabular}{lc}
\hline
\textbf{Model}    & \textbf{Success rate ($\uparrow$)} \\ \hline
ViNT (original)   & 0.44 \\
ViNT (fine-tuned) & 0.89 \\
NoMaD (original)  & 0.10 \\
NoMaD (fine-tuned)& 0.33 \\ \hline
\end{tabular}
\end{table}

Both models exhibit a substantial and consistent improvement after fine-tuning on EgoWalk. 
For NoMaD, we were unable to obtain strong
absolute performance under our particular hardware setup; nevertheless, fine-tuning on EgoWalk more than tripled the success rate relative to the
released checkpoint. 
ViNT performed reasonably well out of the box and
reached a success rate of $0.89$ after fine-tuning. These results support two observations relevant to the broader scaling discussion of end-to-end navigation models. 
First, in the current data regime, simply expanding the pool of training trajectories yields large gains, which underscores the value of contributing new and diverse navigation datasets. 
Second, human-recorded egocentric walking data --- despite the clear embodiment gap --- remains a useful training signal for robot navigation policies.

\subsection{Traversability Segmentation Models}\label{sec:case_traversability}

To demonstrate a downstream use case, we distill SAM's segmentation capability into smaller models for traversability prediction. Specifically, we train a set of lightweight segmentation models in a supervised manner using the \textit{area}-based masks as ground truth. Once trained, these models operate as standard prompt-free segmentation networks, requiring no user-provided prompts at inference time.

We select several model configurations available in the \textit{Segmentation Models PyTorch} \cite{Iakubovskii:2019} package:
\begin{itemize}
    \item SegFormer \cite{xie2021segformer};
    \item U-net \cite{ronneberger2015u} with EfficientNet \cite{tan2019efficientnet} backbone; 
    \item U-Net++ \cite{zhou2018unet++} with ResNet \cite{he2016deep} backbone;
    \item DeepLabV3+ \cite{chen2018encoder} with EfficientNet \cite{tan2019efficientnet} backbone;
    \item FPN \cite{kirillov2017unified} with ResNet \cite{he2016deep} backbone.
\end{itemize}
For all models, the following training setup was employed:
\begin{itemize}
    \item Nvidia A100 GPU (to train multiple models in parallel);
    \item Batch size: 32;
    \item Epochs: 50;
    \item Optimizer: Optimizer: Adam lr=2e-4; cosine annealing scheduler;
    \item Train/val/test split: 85.5/9.5/5\%;
    \item Training took from 6 to 10 hours, depending on the model.
\end{itemize}

Table \ref{tab:segmentation_metrics_results} demonstrates the resulting metrics for these models. It can be seen that all models provide comparable and reasonable results. The smaller models even manage to outperform the larger ones, which is an important fact, since the target robotic platforms usually have limited computational resources. 
Qualitative analysis, provided in Figure \ref{fig:traverse_models_examples}, which reflects these results, showing similar predictions for all models in various environmental conditions.

\begin{table*}
  \scriptsize 
  \setlength{\tabcolsep}{3pt} 
  \caption{Quantitative comparison of different segmentation models' results}
  \label{tab:segmentation_metrics_results}
  \centering
  \begin{tabular}{cccccccc} 
    \toprule
    \textbf{Architecture} & \textbf{Encoder} & \textbf{\makecell{ \# of \\ parameters}} & \multicolumn{5}{c}{\textbf{Metrics}} \\
    & & & \textbf{IoU} & \textbf{F1} & \textbf{Precision} & \textbf{Accuracy} & \textbf{Recall} \\
    \midrule
    Segformer & mit\_b1 & 13M & 0.9062 & 0.9508 & 0.9375 & 0.9645 & 0.9726 \\ 
    Unet & timm-efficientnet-b1 & 6M & \textbf{0.9265} & \textbf{0.9618} & \textbf{0.9542} & 0.9718 & 0.9782 \\
    DeepLabV3+ & efficientnet-b1 & 6M & 0.9252 & 0.9611 & 0.9498 & \textbf{0.9727} & \textbf{0.9784} \\ 
    FPN & se\_resnet50 & 26M & 0.9066 & 0.9510 & 0.9379 & 0.9645 & 0.9727 \\ 
    Unet++ & resnet50 & 23M & 0.8872 & 0.9402 & 0.9214 & 0.9598 & 0.9665 \\ 
    \bottomrule
  \end{tabular}
\end{table*}

\begin{figure}[t]
  \centering
  \includegraphics[width=\linewidth]{figures/photos/traversability_examples_prediction.drawio-compressed.pdf}
  \caption{Comparison between the different segmentation models' capabilities}
  \label{fig:traverse_models_examples}
\end{figure}

\subsection{Language Annotations Evaluation}\label{sec:case_language}
We conduct an expert evaluation to assess the quality of both language-annotation pipelines described in section~\ref{sec:lang_annotation}. For each
pipeline, expert annotators classified a random subset of the samples into five categories: \textit{All Good}, \textit{Partially
Good Caption}, \textit{Bad Caption}, \textit{Bad Goal Selection}, and \textit{All Bad}. Table~\ref{tab:lang_eval} summarizes the results.

\begin{figure*}[!t] 
    \centering
    \includegraphics[scale=0.33]{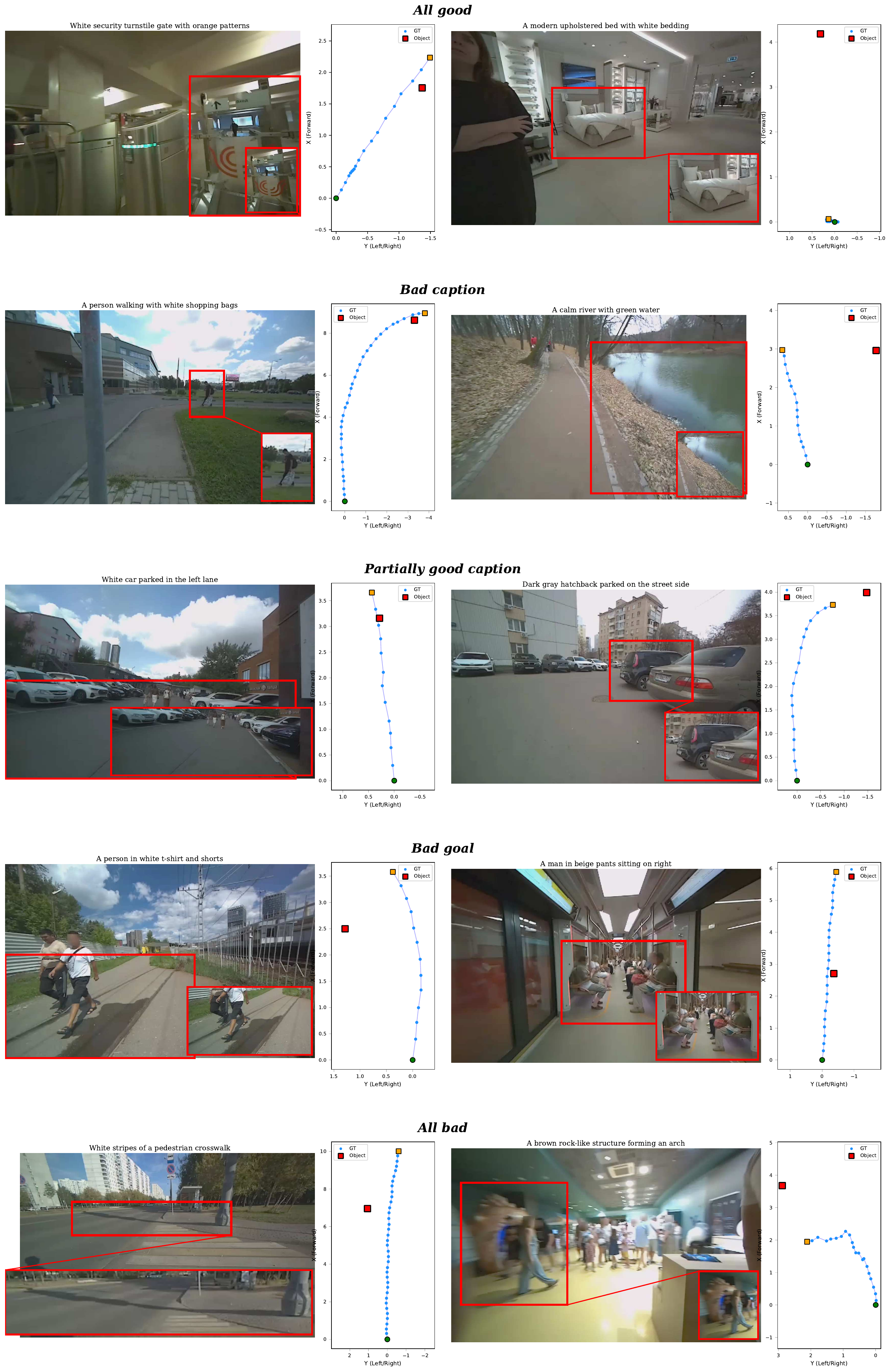} 
    \caption{Qualitative results from language annotations evaluation.}
    \label{fig:lang_qual_results}
\end{figure*}

The \textit{End-to-End} pipeline produces high-quality annotations, confirming that modern VLMs can reliably identify and describe navigation goals when provided with short video context.
The \textit{Goal-Boxes} pipeline performs well but has a notable limitation: its errors are dominated by the geometric goal-selection heuristic, which accounts for 5.3\% of the samples.  
The caption quality is high for both pipelines when the goal is correctly selected, indicating that VLM-based captioning is not the bottleneck.
The two pipelines are thus complementary. \textit{Goal-Boxes} offers tighter spatial grounding through metric back-projection, while
\textit{End-to-End} provides more robust goal identification at the cost of explicit localization. 
Qualitative examples are provided in
Figure \ref{fig:lang_qual_results}.



\begin{table}[t]
  \centering
  \caption{Quality evaluation of the two language-annotation pipelines. Expert annotators classified random samples into five categories.}
  \label{tab:lang_eval}
  \begin{tabular}{lcccc}
    \toprule
    & \multicolumn{2}{c}{\textbf{Goal-Boxes}} & \multicolumn{2}{c}{\textbf{End-to-End}} \\
    \cmidrule(lr){2-3} \cmidrule(lr){4-5}
    \textbf{Category} & \# & \% & \# & \% \\
    \midrule
    All Good              & 647 & 87.7 & 740 & 99.1 \\
    Partially Good Caption&  12 &  1.6 &   1 &  0.1 \\
    Bad Caption           &  22 &  3.0 &   3 &  0.4 \\
    Bad Goal Selection    &  39 &  5.3 &   0 &  0.0 \\
    All Bad               &  18 &  2.4 &   3 &  0.4 \\
    \midrule
    \textit{Samples evaluated} & \multicolumn{2}{c}{738} & \multicolumn{2}{c}{747} \\
    \midrule
    \textbf{Goal Selection Acc.}    & \multicolumn{2}{c}{92.3} & \multicolumn{2}{c}{99.6} \\
    \textbf{Caption Quality ($\geq$ partial)} & \multicolumn{2}{c}{89.3} & \multicolumn{2}{c}{99.2} \\
    \bottomrule
  \end{tabular}
\end{table}

\subsection{Reconstruction and Pose Estimation Demo}

\begin{figure}[h!]
    \centering
    \includegraphics[height=3.2cm, width=0.75\linewidth]{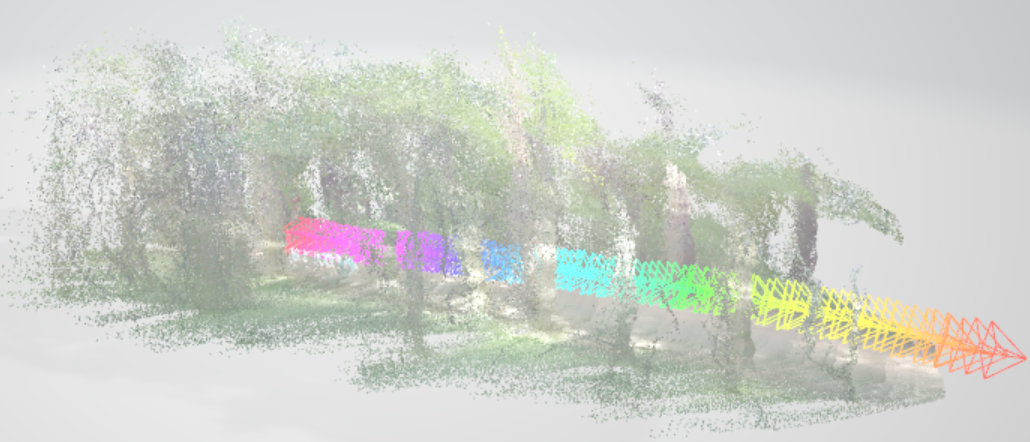}
    \caption{Before gaussian splatting, the EgoWalk scene is reconstructed by estimating depth and camera poses.}
    \label{fig:pcd_forest}
\end{figure}

To illustrate the potential of the EgoWalk dataset for real2sim pipelines, we employ a reconstruction, pose estimation, and novel view synthesis (NVS) pipeline built upon the recent Depth Anything 3 (DA3) family of models~\cite{lin2025depth}. Given a set of $N$ input images, DA3 produces $N$ pixel-aligned depth and ray maps, which are subsequently fused into point clouds (see Figure~\ref{fig:pcd_forest} for an example from EgoWalk), yielding high-fidelity 3D Gaussians and geometry. Notably, this approach recovers spatially consistent geometry even when camera poses are unknown.

\begin{figure}[h!]
    \centering
    \begin{subfigure}[b]{0.8\linewidth}
        \centering
        \includegraphics[height=2.2cm, width=\linewidth]{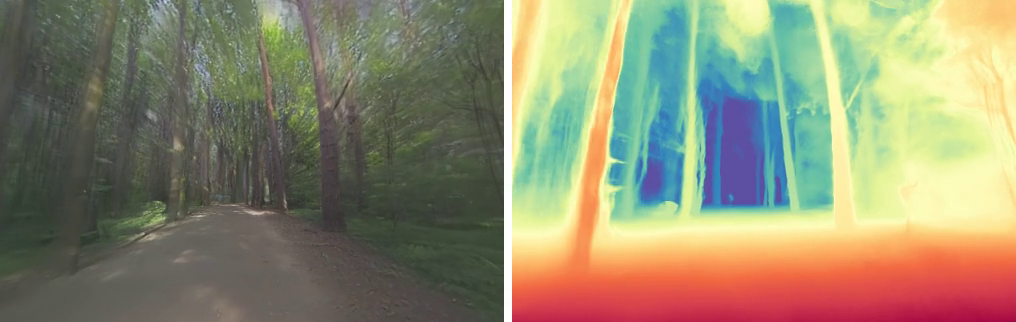}
    \end{subfigure}
    
    \vspace{1em} 
    
    \begin{subfigure}[b]{0.8\linewidth}
        \centering
        \includegraphics[height=2.2cm, width=\linewidth]{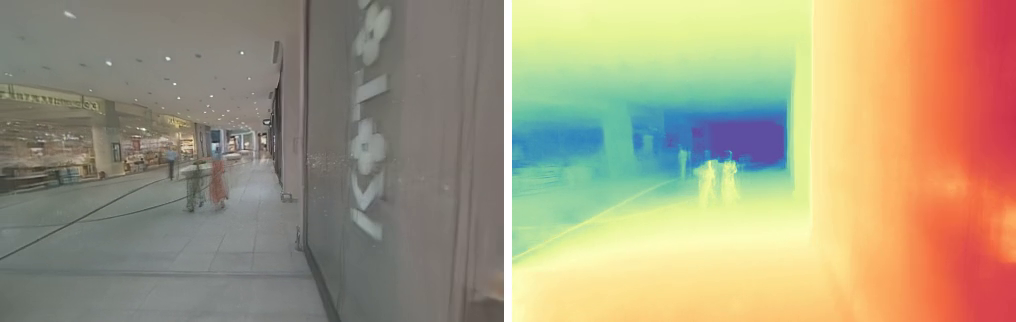}
    \end{subfigure}
    \caption{After reconstruction, we can render images according to defining poses, which is important for robot interaction and navigation.}
    \label{fig:scene_reconstruction_rendering_pose}
\end{figure}

\begin{figure}[h!]
    \centering
    \begin{subfigure}[b]{0.4\linewidth}
        \centering
        \includegraphics[height=2.2cm, width=\linewidth]{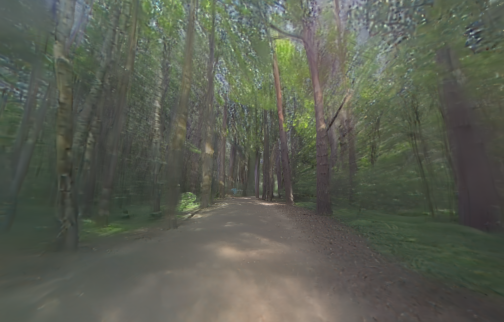}
        \label{fig:rendered_frame_pred_0000_col1}
    \end{subfigure}
    \hspace{-0.6em}
    \begin{subfigure}[b]{0.4\linewidth}
        \centering
        \includegraphics[height=2.2cm, width=\linewidth]{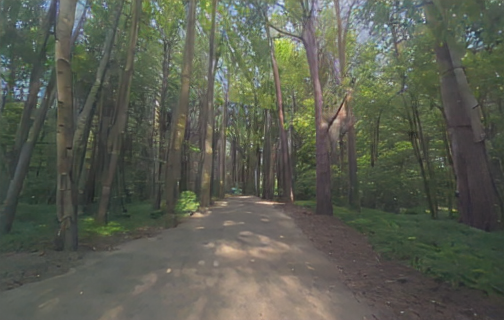}
        \label{fig:restored_rendered_frame_pred_0000_col2}
    \end{subfigure}
    
    \vspace{-0.1em} 

    \begin{subfigure}[b]{0.4\linewidth}
        \centering
        \includegraphics[height=2.2cm, width=\linewidth]{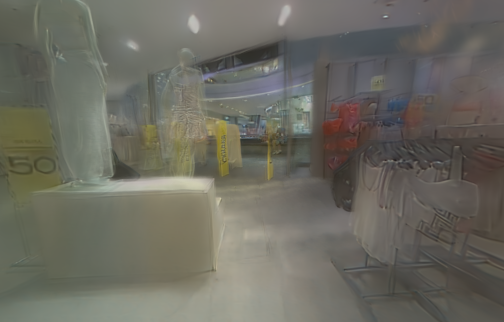}
        \label{fig:rendered_frame_pred_0000_col1}
    \end{subfigure}
    \hspace{-0.6em}
    \begin{subfigure}[b]{0.4\linewidth}
        \centering
        \includegraphics[height=2.2cm, width=\linewidth]{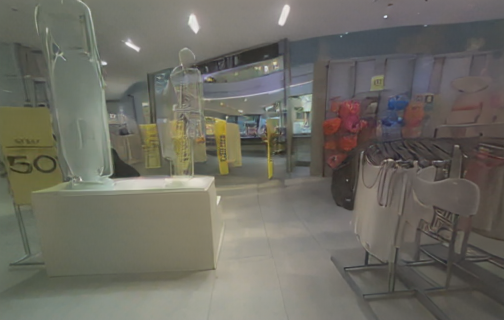}
        \label{fig:restored_rendered_frame_pred_0000_col2}
    \end{subfigure}
    \caption{Rendered images with artifacts after Gaussian Splatting (right side) and improved images (left side) using Difix3D+ to reduce the noise.}
    \label{fig:Difix3D_improved}
\end{figure}

Once reconstruction is complete, novel views can be rendered using DA3's built-in Gaussian splatting (GS) module (see Figure~\ref{fig:scene_reconstruction_rendering_pose}). Although GS enables photorealistic rendering, characteristic GS artifacts frequently remain in
the synthesized images. Recent enhancement models, such as Difix3D+~\cite{wu2025difix3d+}, are designed to mitigate these artifacts and improve rendering quality, as illustrated in Figure~\ref{fig:Difix3D_improved}.

To provide an initial assessment of DA3's reconstruction capabilities, we
constructed a pose estimation benchmark on top of EgoWalk. We selected
60 sub-trajectories from moderate and challenging scenes, each consisting
of 100 frames sampled at the standard frame rate (i.e. 5 FPS), and treated the poses
provided by EgoWalk as ground truth. Because these reference poses are
obtained through visual-inertial odometry, the benchmark cannot be regarded
as a definitive standard; nevertheless, it serves as a useful proxy for
reconstruction quality, which is tightly coupled to pose estimation accuracy. Several variants of the DA3 model were evaluated on this benchmark.

We adopt two standard metrics \cite{xu2024accurate}: the Relative Rotation Accuracy (RRA),
based on the angular error $e_R$ between the predicted and
ground-truth relative rotations, and the Relative Translation Accuracy
(RTA), based on the angular error $e_t$ between the predicted and
ground-truth translation directions. We summarize both into a single per-pair error $e = \max(e_R, e_t)$ and seek to keep it small. For a
test set of $N$ images and a threshold $\tau$ (in degrees), the
accuracy–threshold curve is defined as
\begin{equation}
    \operatorname{Acc}(\tau) = \frac{1}{N} \sum_{i=1}^{N}
    \mathbf{1}\!\left(e_i \leq \tau\right),
\end{equation}
and the corresponding area under the curve (AUC) is
\begin{equation}
    \mathrm{AUC}_{\tau_{\max}} = \frac{1}{\tau_{\max}}
    \int_{0}^{\tau_{\max}} \operatorname{Acc}(\tau)\, d\tau,
\end{equation}
where $\mathrm{AUC}@3$ reflects a strict accuracy regime and
$\mathrm{AUC}@30$ a more permissive one~\cite{lin2025depth}. The pose estimation results are reported in Table~\ref{tab:da3_auc_metrics}.

\begin{table}[htbp]
\centering
\caption{AUC Performance Metrics for DA3 Model Variants}
\label{tab:da3_auc_metrics}
\begin{tabular}{lcccc}
\toprule
\textbf{Model} & \textbf{AUC@3} & \textbf{AUC@5} & \textbf{AUC@15} & \textbf{AUC@30} \\
\midrule
DA3-Giant & 22.82 & 38.30 & 72.65 & 85.15 \\
DA3-Large &  9.49 & 20.44 & 59.48 & 77.96 \\
DA3-Base  &  5.98 & 13.61 & 43.21 & 63.71 \\
DA3-Small &  3.49 &  8.68 & 33.72 & 53.98 \\
\bottomrule
\end{tabular}
\end{table}

Pose estimation metrics are compared between different DA3 model versions. As it was expected, the bigger model showed better results on our unseen data. We refer to the original table in paper~\cite{lin2025depth}, where the models are estimated on easier scenarios. By comparing two tables, you can see that our difficult environments demonstrated the possibility for future improvements of pose estimation methods.     

\begin{figure}[H]
    \centering
    \begin{subfigure}[b]{0.75\linewidth}
        \centering
        \includegraphics[height=2.7cm, width=\linewidth]{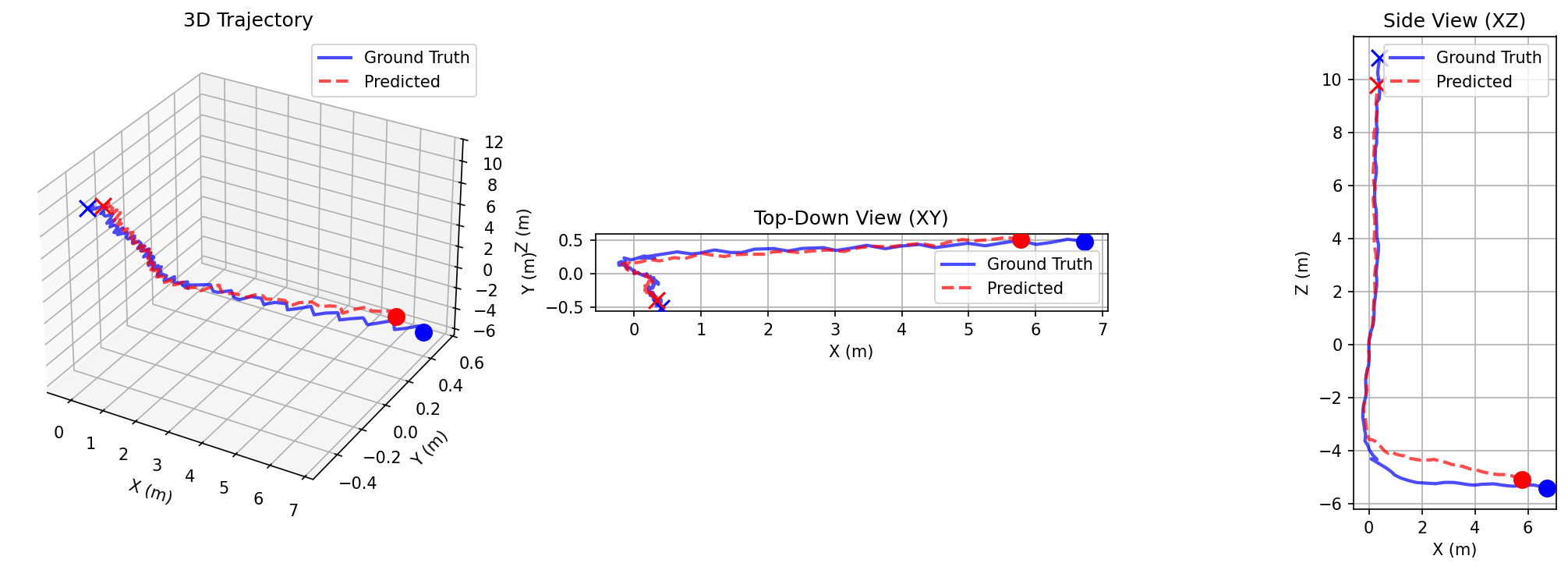}
    \end{subfigure}
    \begin{subfigure}[b]{0.75\linewidth}
        \centering
        \includegraphics[height=2.7cm, width=\linewidth]{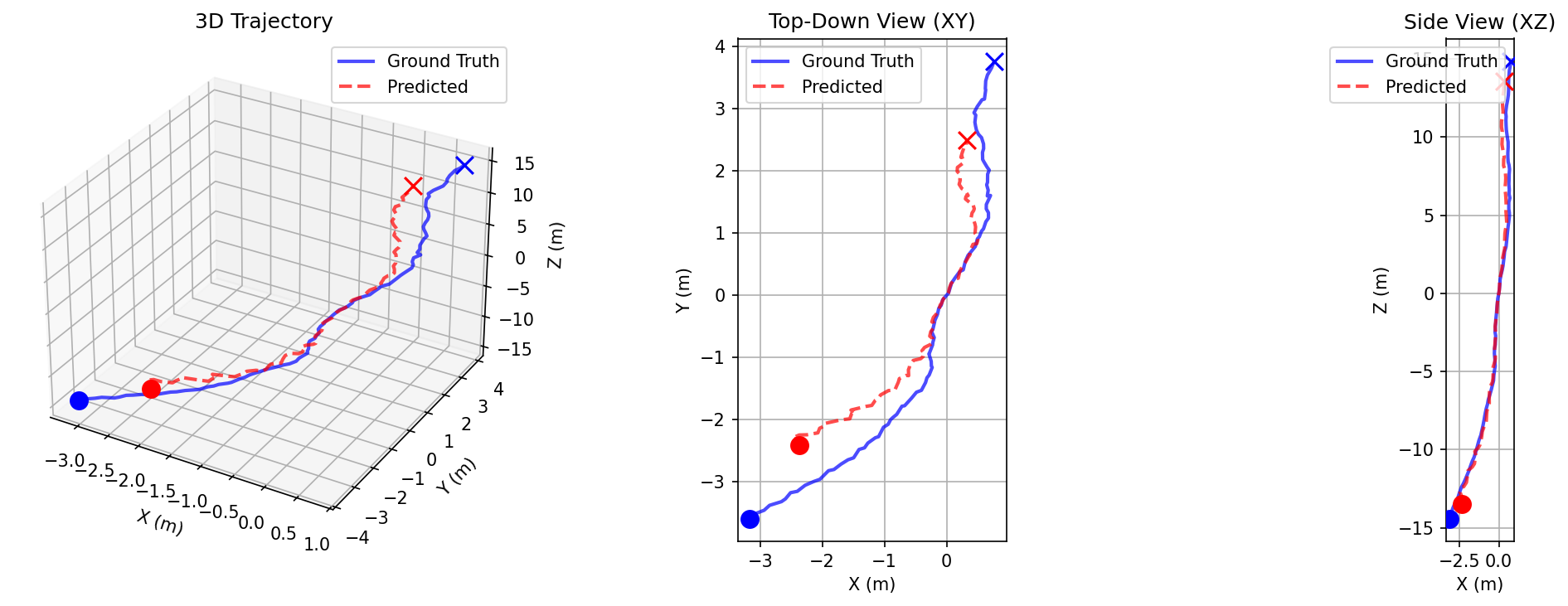}
    \end{subfigure}
    \caption{Estimated trajectories for the outdoor and indoor scenes, which are shown in Figure \ref{fig:Difix3D_improved}.}
    \label{fig:trajectory_estimation}
\end{figure}

Scene reconstruction and trajectory estimation are essential tasks for robot-human interaction and navigation. Although there are many methods to solve such tasks, there is a high demand for various data obtained in different conditions to further improve and increase the robustness of current methods according to Table \ref{tab:da3_auc_metrics} and Figure \ref{fig:trajectory_estimation}. We showed that there is still some potential to improve navigation methods.

\section{Limitations} \label{sec:limitations}

Despite offering a diverse and richly annotated dataset, our work comes with several important limitations. While recording volunteers were carefully instructed, human motion is inherently noisy and non-deterministic, since the humans are noisy rational \cite{kwon2020humans} agents by nature. As a result, the recorded trajectories may exhibit behaviors such as near-collisions or maneuvers that are infeasible for robotic platforms. In addition, although we employ advanced visual-inertial odometry provided by the \textit{ZED SDK}, the resulting pose estimates are not flawless. We observed occasional failures in challenging conditions, particularly in low-light environments, which may affect trajectory precision. Future improvements in odometry methods could help address these issues.

Furthermore, our annotation pipelines—for example, those generating natural language goals and traversability masks—are based on heuristics. While scalable and efficient, they may yield suboptimal or incorrect outputs in edge cases. These limitations could potentially be mitigated through more sophisticated combinations of large language models (LLMs), vision-language models (VLMs), and pre-trained 2D/3D detection and segmentation networks, which we identify as promising directions for future work.

The real-world navigation evaluation reported in
Section \ref{sec:vn_real_world_eval} is necessarily limited in
statistical power. Resource constraints prevented us from running
experiments across a larger number of environments — in particular,
outdoor ones — or from conducting more trials per condition.
Nevertheless, we observed that the navigation policies behaved in a
largely deterministic manner: the trajectories produced across repeated
trials in the same environment exhibited little variation, which
suggests that even a small number of trials is sufficient for an initial
assessment of the impact of new training data.

Societal impact of recording a dataset in common human spaces provides positive examples of navigation in environments where people typically transition. This dataset subtly captures the interactions with other pedestrians while the volunteer is navigating, and it provides examples for training robot navigation policies that adhere to the common rules of navigating in such environments. As a negative impact, personal data could be leaked from the people appearing in the dataset, perhaps not their faces (blurred), but other kind of information about body, nearby objects, etc. We believe the benefits outweigh the risks.

\section{Conclusions}

Our work introduces a novel, large-scale, and diverse dataset for visual navigation tasks to support research in goal-conditioned policy learning, scene understanding, and language grounding. We release the dataset along with open-source code for data processing, annotation, and benchmark evaluation to promote reproducibility and further research.

Automatic data extraction, processing, and annotation approaches are introduced, along with the anticipated practical applications. 
Qualitative and quantitative evaluations, including real robot experiments, outlined important insights on existing bottlenecks in navigation policies and requirements for future dataset collections. Important limitations of the provided data are also outlined, such as human motion noise, odometry errors, and the heuristic nature of annotations. Future work will focus on addressing these limitations, providing informative formal benchmarks for the quality of both data and resulting navigation policies, and enriching the dataset with new recordings and carefully curated annotations.

\section*{Acknowledgments}
The work was supported by the grant for research centers in the field of AI provided by the Ministry of Economic Development of the Russian Federation in accordance with the agreement 000000C313925P4F0002 and the agreement with Skoltech №139-10-2025-033.

The authors would like to acknowledge the use of AI-assisted tools in the preparation of this manuscript. Specifically, Claude \cite{anthropic2025}, Gemini 3 \cite{google2025gemini3} and Deep Writer \cite{deepwriter2025} were used to assist with grammar checking and text logic improvement. The authors take full responsibility for the accuracy and integrity of the content presented in this work.



\bibliographystyle{plain}
\bibliography{ref}

\begin{IEEEbiography}[{\includegraphics[width=1in,height=1.25in,clip,keepaspectratio]{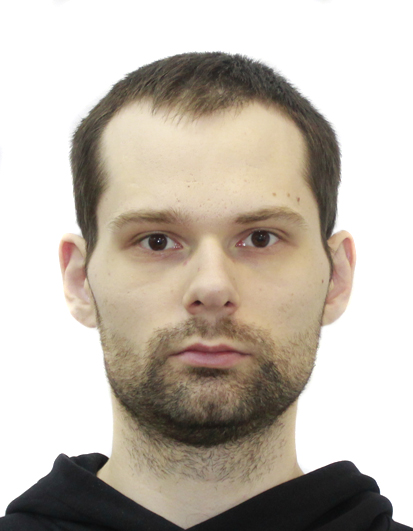}}]{Timur Akhtyamov} received the B.Sc. degree from Bauman Moscow State Technical University and the M.Sc. degree from the Skolkovo Institute of Science and Technology. His background encompasses software engineering, machine learning, and robotics, with a primary focus on autonomous navigation. He is currently pursuing the Ph.D. degree, with his thesis focusing on social robot navigation utilizing optimization and learning techniques.
\end{IEEEbiography}

\begin{IEEEbiography}[{\includegraphics[width=1.1in,height=1.75in,clip,keepaspectratio]{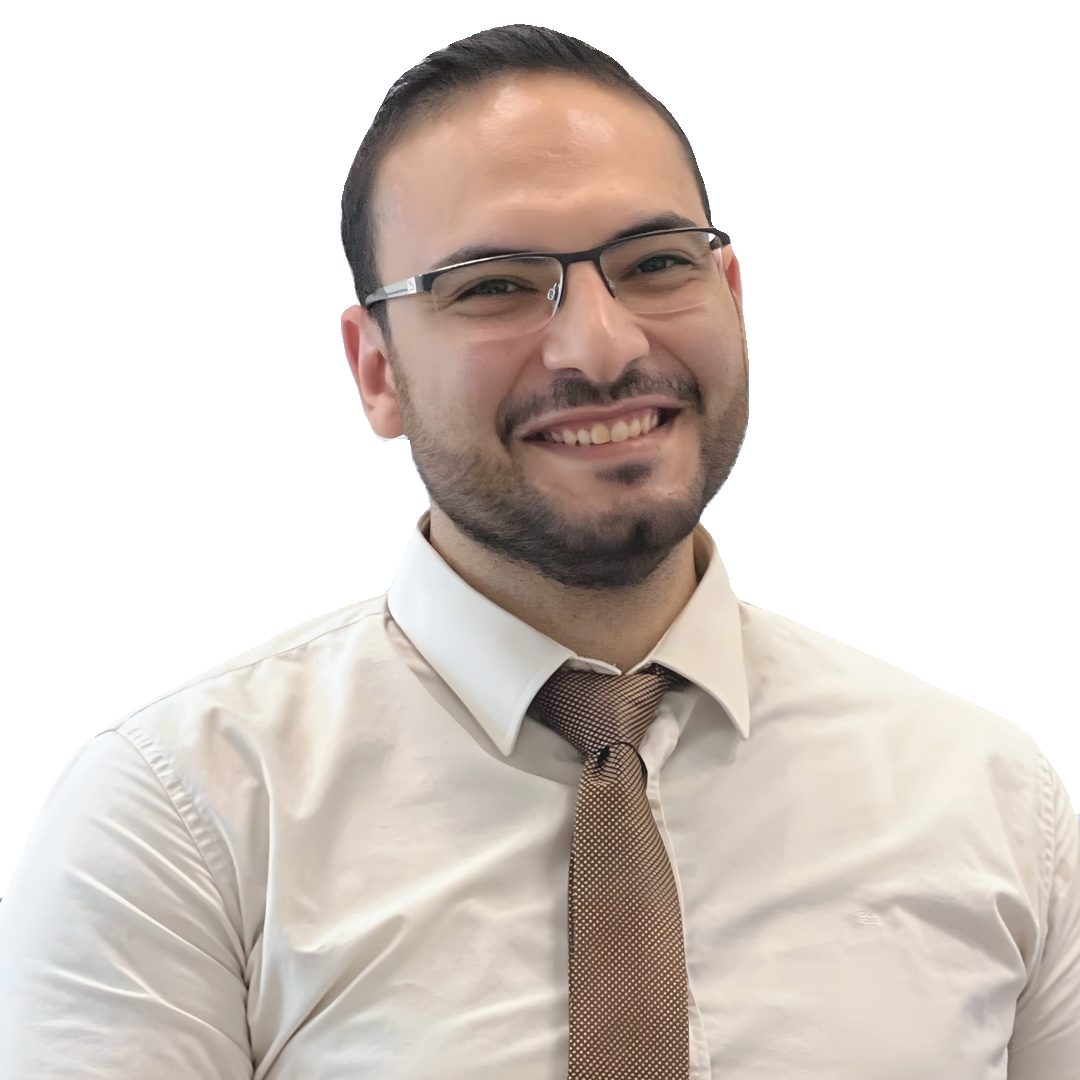}}]{Mohamad Al Mdfaa} received the B.Eng. degree in Computer and Automation Engineering from Damascus University, in 2017, and M.Sc. degree in Robotics and Computer Vision from Innopolis University, in 2022. He is currently pursuing the Ph.D. degree in robotics at Skolkovo Institute of Science and Technology (Skoltech). His research interests include deep learning, robotics, scene graph representations, LLM, VLM, and multi-agent AI systems.
\end{IEEEbiography}

\begin{IEEEbiography}[{\includegraphics[width=1.1in,height=1.75in,clip,keepaspectratio]{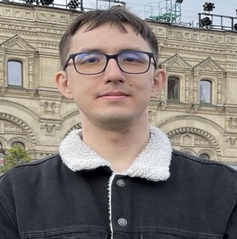}}]{Nigmatzyanov Arthur Rashidovich} received the B.Sc. and M.Sc. degrees in computer science at Moscow Institute of Physics and Technology. Since 2022 he has been pursuing a Ph.D. at Skolkovo Institute of Science and Technology. His research interests: 3D computer vision, remote sensing, point clouds, reconstruction, and foundation models.
\end{IEEEbiography}

\begin{IEEEbiography}[{\includegraphics[width=1.1in,height=1.75in,clip,keepaspectratio]{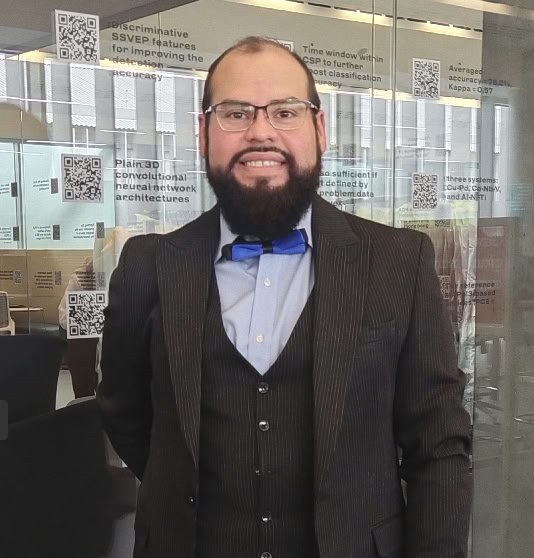}}]{Javier Antonio Ramírez Benavides} is a research engineer specializing in nanomaterials, applied physics, and mobile robotics, currently based in Moscow. He received his Ph.D. in Materials Science and Engineering from the Skolkovo Institute of Science and Technology, focusing on advanced carbon nanomaterials for optoelectromechanical applications . He also holds an M.Sc. in Applied Physics and Mathematics from the Moscow Institute of Physics and Technology and a B.Eng. in Information Technologies from IUTFRP, Venezuela.

His research spans nanomaterial synthesis, spectroscopy, photonics, and plasma systems, with contributions to carbon-based nanomaterials and a patented filtration membrane technology. He currently works in mobile robotics, focusing on system integration, autonomous platforms, and robotic infrastructure.
\end{IEEEbiography}

\begin{IEEEbiography}[{\includegraphics[width=1in,height=1.25in,clip,keepaspectratio]{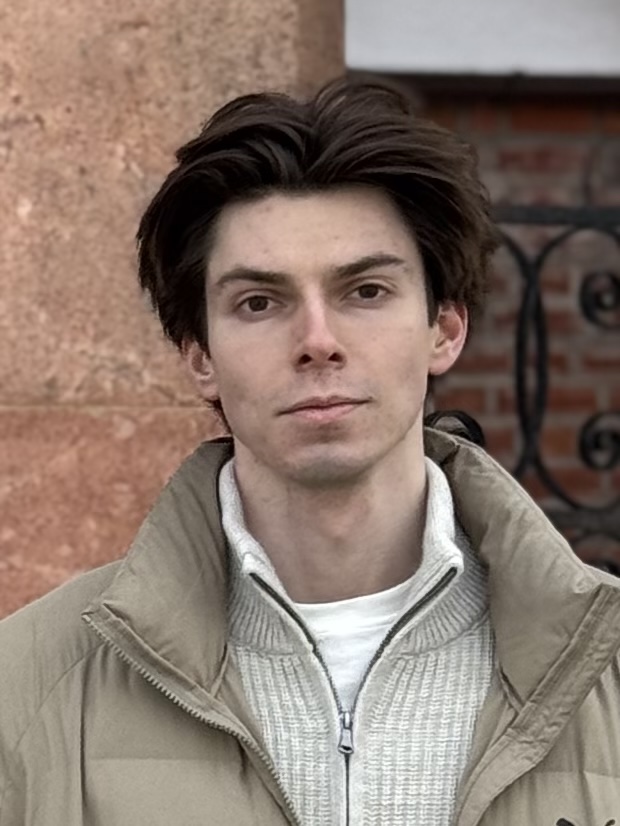}}]{Sergey Bakulin} received the B.Sc. degree in Underwater Robotics from Bauman Moscow State Technical University in 2023 and the M.Sc. degree in Engineering Systems from Skolkovo Institute of Science and Technology, where he is currently pursuing a Ph.D. in Computational and Data Science and Engineering.

He is a researcher at the Sber Robotics Center. His research focuses on reinforcement learning for robotics, robot navigation, and learning-based perception, with an emphasis on representation learning and spatial encoding.

His current interests include humanoid locomotion, navigation, learning-based representations for robotics and Vision-Language-Action (VLA) models.
\end{IEEEbiography}

\begin{IEEEbiography}
[{\includegraphics[width=1in,height=1.25in,clip,keepaspectratio]{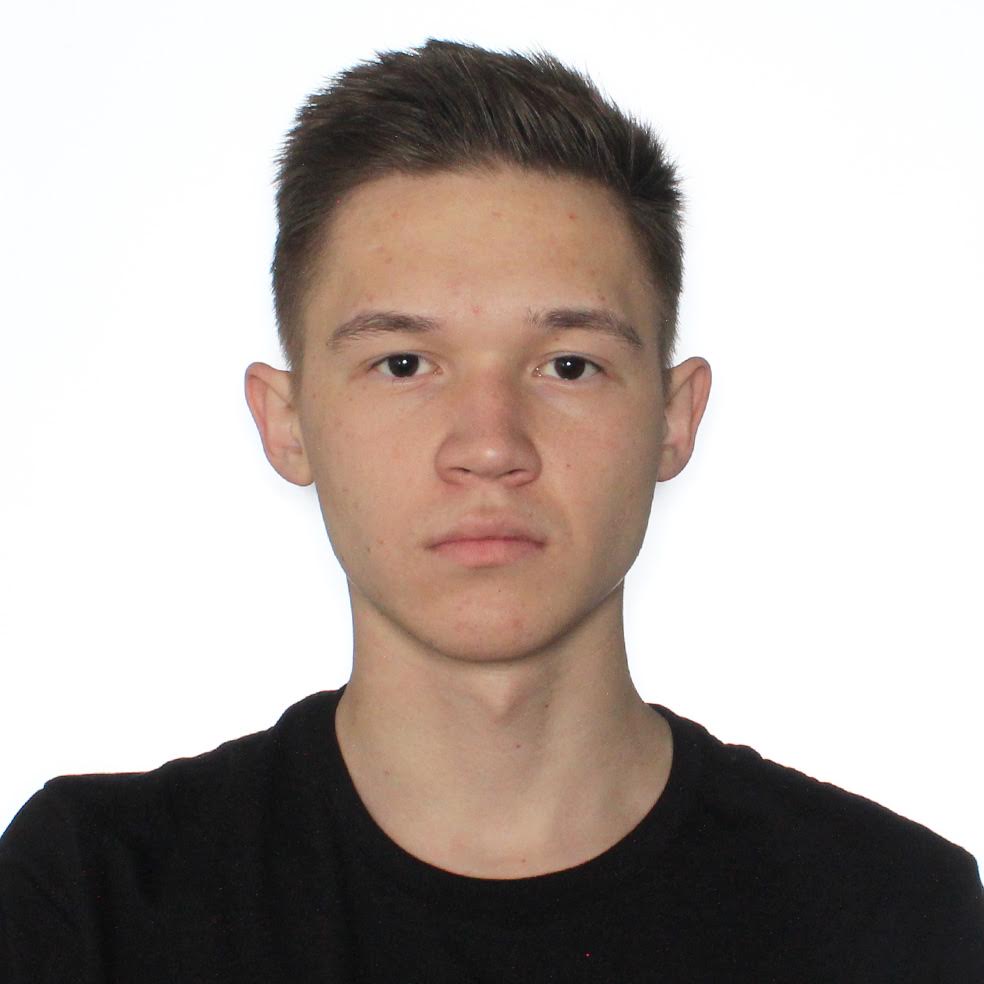}}]{German Devchich} received the B.S. degree in automation of technological processes and production from Bauman Moscow State Technical University, Moscow, Russia, in 2024. He is currently pursuing the M.S. degree in data science with the Mobile Robotics Lab, Skolkovo Institute of Science and Technology (Skoltech), Moscow, Russia, where he is also a Research Scientist. His research interests include 3D Gaussian splatting, autonomous driving simulation, pedestrian detection, computer vision, and LiDAR-based perception.
\end{IEEEbiography}

\begin{IEEEbiography}
[{\includegraphics[width=1in,height=1.25in,clip,keepaspectratio]{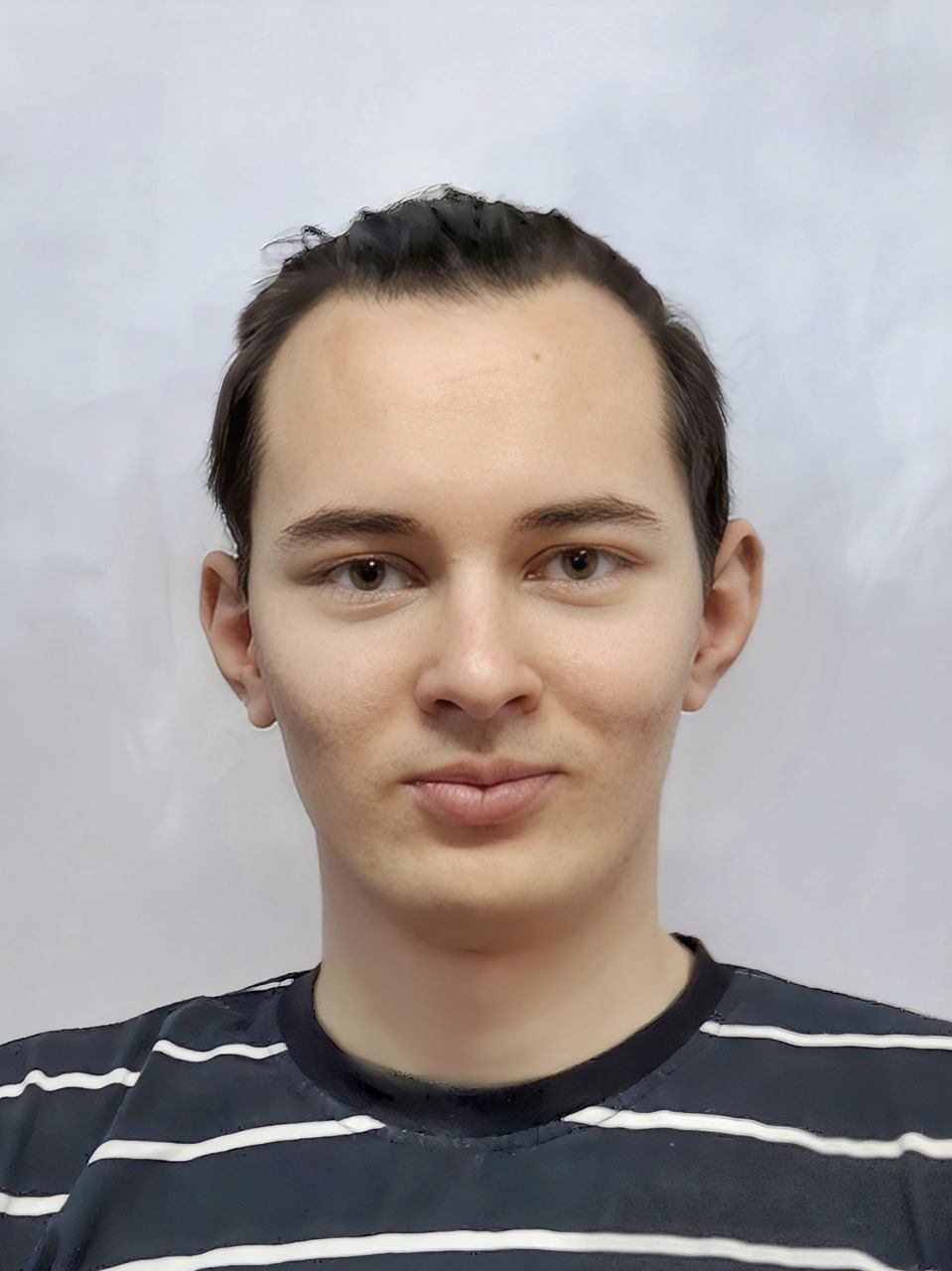}}]{Denis Fatykhoph} received the B.Sc. degree in Robotics from Bauman Moscow State Technical University in 2024, and is currently pursuing the M.Sc. degree in Data Science at the Applied AI Center, Moscow, Russia. His current research interests include visual localization and topometric navigation, with a focus on segment-level matching for semantic navigation graphs using foundation models. He is also involved in projects related to visual SLAM, 3D perception, and autonomous robot navigation.
\end{IEEEbiography}

\begin{IEEEbiography}
[{\includegraphics[width=1in,height=1.25in,clip,keepaspectratio]{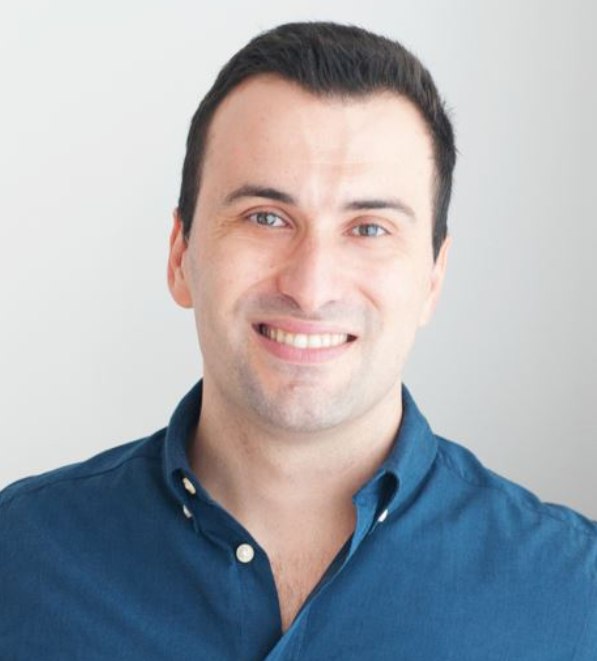}}]{Diego Ruiz Salinas} is a Research Collaborator at Skoltech, focusing on AI-driven Robot Navigation. He holds an M.Sc. in Renewable Energy Engineering from University of Applied Sciences Technikum Wien. Before pivoting to academic research, Diego built an extensive career in industry as an AI Engineer at Atos, Avanade, and Eugen, specializing in neural networks and machine learning systems. He is currently focused on bridging the gap between advanced AI engineering and autonomous robotic perception.
\end{IEEEbiography}

\begin{IEEEbiography}
[{\includegraphics[width=1in,height=1.25in,clip,keepaspectratio]{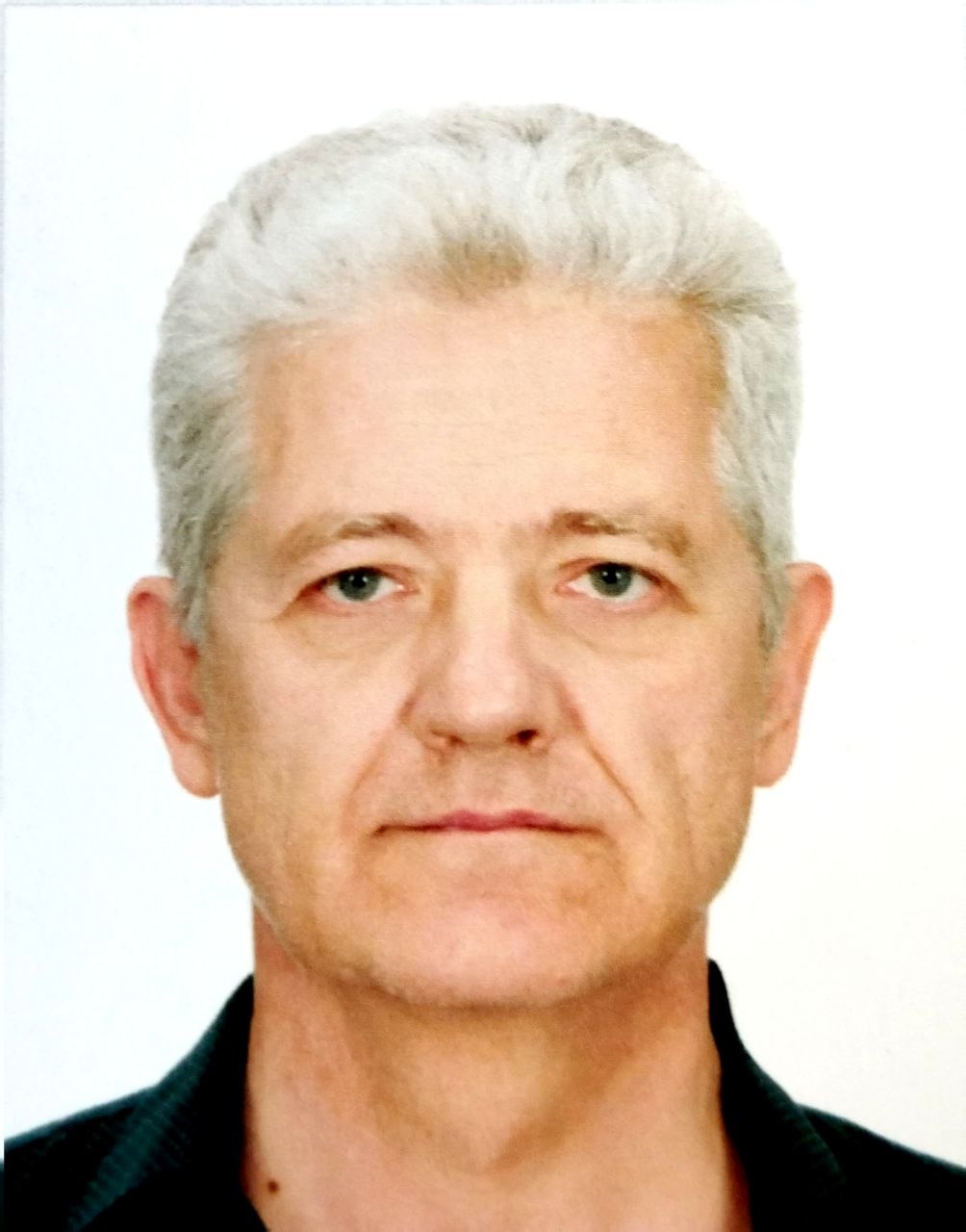}}]{Alexander Mazurov} received the M.Sc. degree from the Ural State University of Economics, specializing in data processing and management. He previously worked in the media departments of the Sverdlovsk Oblast. He is currently based in Moscow, where he contributes to open-source data mining projects and is actively involved in cinema and theater.
\end{IEEEbiography}

\begin{IEEEbiography}
[{\includegraphics[width=1in,height=1.25in,clip,keepaspectratio]{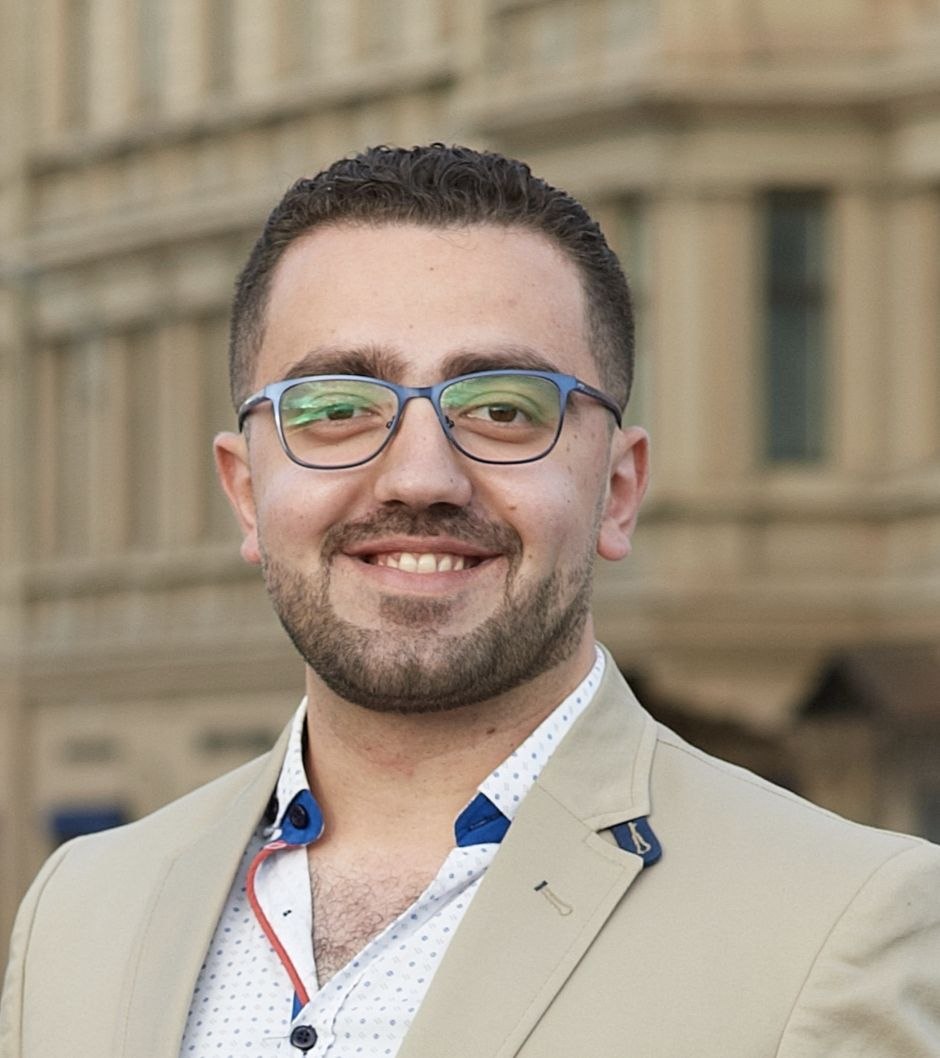}}]{Malik Mohrat} received the B.S. degree in mechatronics engineering from Homs University and the M.S. degree in robotics from ITMO University, where he is currently pursuing the Ph.D. degree in mobile robotic systems. His research focuses on neural network methods for metric-semantic 3D mapping, including 3D Gaussian Splatting and its integration with SLAM, as well as object-centric 3D scene representation. His research interests include computer vision, machine learning for robotics, and autonomous systems. He is also involved in the development of humanoid robotic systems at the Robotics Center in Moscow
\end{IEEEbiography}

\begin{IEEEbiography}
[{\includegraphics[width=1in,height=1.25in,clip,keepaspectratio]{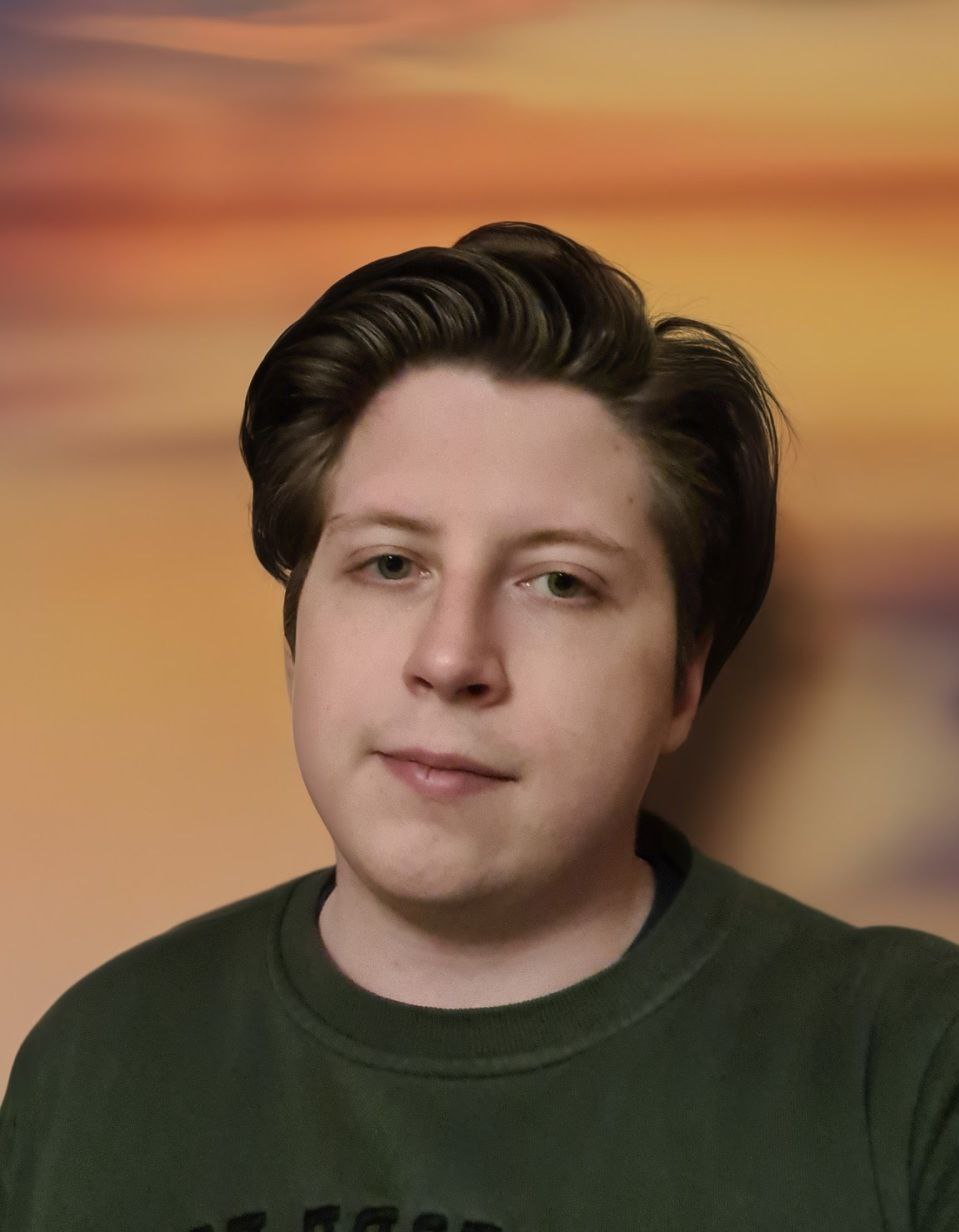}}]{Pavel Kolesnik} received the M.Sc. degree in Computer Systems and Computer Networks at the Higher School of Economics, Moscow, Russia, in 2022. His early research focused on hand prosthesis development and EEG-based control systems for prosthetic devices.
He is currently a researcher at the Robotics Center, Moscow, Russia, where he works on the development and application of artificial intelligence methods for advanced robotic navigation and autonomous systems.
His research interests include deep learning, robotics, large language models (LLM), vision-language models (VLM), agentic systems, reinforcement learning, world models, and Vision-Language-Action (VLA) models
\end{IEEEbiography}

\begin{IEEEbiography}
[{\includegraphics[width=1in,height=1.25in,clip,keepaspectratio]{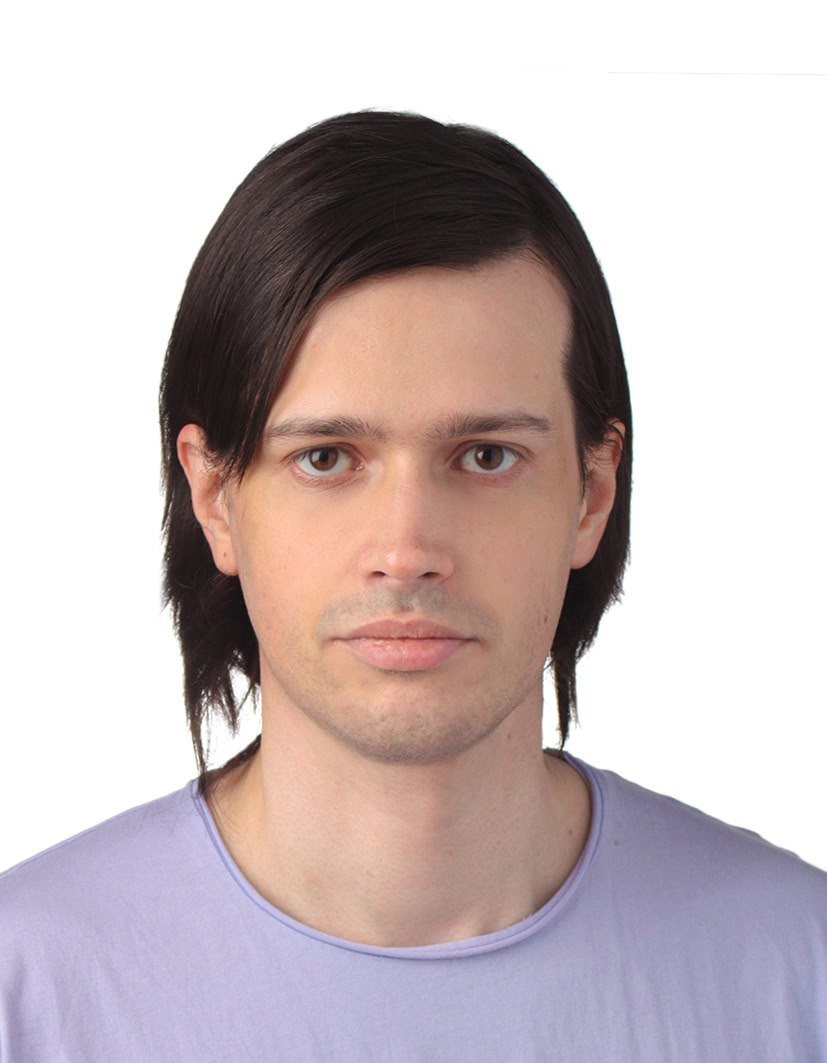}}]{Ivan Sosin} received the M.Sc. degree in Applied Mathematics at the Academic University, Saint Petersburg, Russia, in 2018. His early research focused on sEMG sensor data processing for gesture recognition.

He is currently working at the Robotics Center, Moscow, Russia. His research focuses on the development and application of artificial intelligence for advanced robotic navigation.

His research interests include deep learning, robotics, LLM, VLM, agents, reinforcement learning, world models, Vision-Language-Action (VLA) models.
\end{IEEEbiography}

\begin{IEEEbiography}
[{\includegraphics[width=1in,height=1.25in,clip,keepaspectratio]{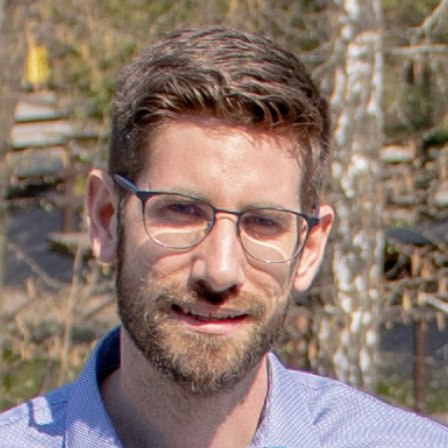}}]{Gonzalo Ferrer} obtained  his Ph.D. in Robotics from the {\em Universitat Polit\`ecnica de Catalunya} (UPC), Barcelona, Spain in 2015 and worked during two years as a Research Fellow (postdoc) at the APRIL lab. in the department of Computer Science and Engineering at the University of Michigan. In 2018, Gonzalo  joined the Skolkovo Institute of Science and Technology as an Assistant Professor. He is heading the Mobile Robotics lab., focusing his research on planning, perception and how to combine both into new solutions in robotics.
\end{IEEEbiography}





\EOD
\end{document}